\documentclass{article}

\usepackage{graphicx}

 \usepackage[eandd, final]{neurips_2026}

\usepackage[utf8]{inputenc} %
\usepackage[T1]{fontenc}    %
\usepackage{hyperref}       %
\usepackage{url}            %
\usepackage{booktabs}       %
\usepackage{amsmath}
\usepackage{amsfonts}       %
\usepackage{nicefrac}       %
\usepackage{microtype}      %
\usepackage{xcolor}         %

\usepackage{hyperref}
\usepackage{url}
\usepackage{graphicx}
\usepackage{fvextra}
\usepackage{tabularx}

\DefineVerbatimEnvironment{PromptBox}{Verbatim}{%
  fontsize=\footnotesize,
  breaklines=true,
  breakafter={|,;,:},
  breaksymbolleft={},
  breaksymbolright={},
  frame=single,
  framerule=0.2pt,
  framesep=2mm
}

\title{Beyond Accuracy: A Diagnostic Benchmark for Hypothesis-Driven Experiment Planning in LLM Agents}

\author{
Ryo Kuroki$^{1,2}$\thanks{Equal contribution}
\qquad Amin Mansouri$^{1}$\footnotemark[1] \qquad Philippe Schwaller$^{1}$ \\
$^{1}$ EPFL, Lausanne, Switzerland \qquad $^{2}$ Asahi Kasei Corporation, Tokyo, Japan \\
}

\hypersetup{
  pdftitle={Beyond Accuracy: A Diagnostic Benchmark for Hypothesis-Driven Experiment Planning in LLM Agents},
  pdfauthor={Ryo Kuroki, Amin Mansouri, Philippe Schwaller}
}

\begin{document}

\maketitle
\setcounter{footnote}{0} %

\begin{abstract}

Evaluating large language model (LLM) agents on scientific tasks typically reduces to final-task accuracy, which obscures where reasoning succeeds or breaks down. We introduce a benchmark and diagnostic evaluation framework for hypothesis-driven adaptive experimental planning in closed-world system identification, instantiated in kinetic mechanism identification. The framework goes beyond final accuracy in two ways. First, we evaluate exploration behavior by quantifying the extent to which agents exhibit space-filling or corner-seeking patterns in the experimental design space. Second, we decompose the reasoning process into four interpretable steps: belief update, hypothesis discrimination, discriminative experiment design, and predictive reasoning under interventions, each scored separately on reasoning traces. Across ten LLM agents, we find that stronger agents outperform both adaptive (Bayesian optimization) and non-adaptive (Latin hypercube) baselines and exhibit exploration patterns inconsistent with naive heuristics. Decomposing reasoning reveals that the stronger agents score consistently well across all four steps, whereas weaker agents often succeed at belief update and hypothesis discrimination but are less reliable at translating them into actionable designs and sharp contrastive predictions.
Independent expert annotations support predictive reasoning under our stated validation criteria, while evidence for discriminative experiment design is positive but less consistent. The results suggest a gap between recognizing plausible hypotheses and translating them into contrastive predictions, with more tentative evidence for a corresponding experiment-design bottleneck. The benchmark website is available at \url{https://schwallergroup.github.io/beyond-accuracy/}.
\end{abstract}

\section{Introduction}
\begin{figure}[t]
    \centering
    \includegraphics[width=1.0\linewidth]{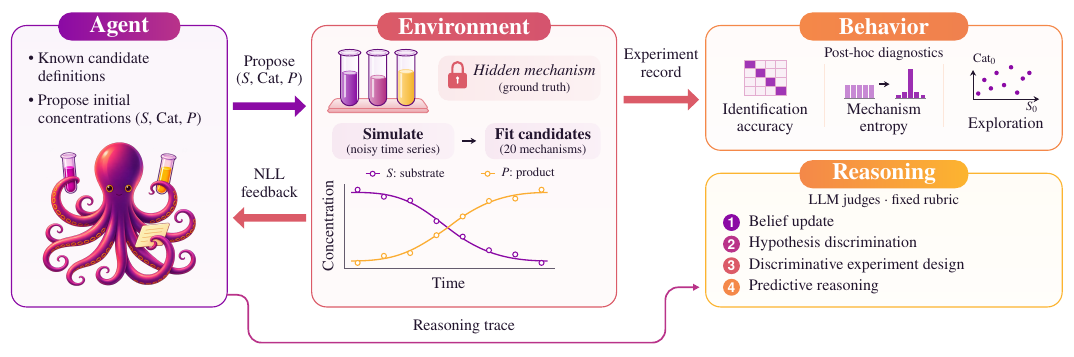}
    \caption{
    Overview of the benchmark and diagnostic evaluation framework. Agents distinguish among 20 candidate mechanisms over five experimental rounds on 200 reaction systems. In the agent--environment loop (left), the agent proposes an experimental condition $\alpha_k$. The environment simulates noisy kinetic time-series observations under a hidden ground-truth mechanism and returns negative log-likelihood (NLL) values obtained by fitting candidate mechanisms to the accumulated observations. The diagnostic framework (right) evaluates mechanism identification accuracy, mechanism entropy, and exploration behavior, and scores four reasoning steps of the agent's reasoning trace using LLM judges. The time-series curves, observation markers, and behavioral mini-plots are schematic.
    }
    \label{fig:overview}
\end{figure}

Identifying physical and chemical systems from time-series observations is a fundamental problem in scientific discovery. 
This class of problems arises in multiple scientific domains where competing dynamical models must be discriminated through adaptive experimentation. Examples include kinetic mechanism identification in chemistry \citep{blackmond2005reaction, bures2016simple}, adaptive Hamiltonian learning in quantum systems \citep{wang2017experimental, wiebe2014hamiltonian}, gene regulatory network inference under perturbation experiments in systems biology \citep{gardner2003inferring, bonneau2006inferelator}.
Researchers typically refine beliefs over candidate models through a sequence of experiments designed to distinguish among them. Here, we study adaptive experiment planning in a controlled closed-world setting with fixed, known candidate mechanisms. This setting allows us to evaluate adaptive experiment planning and diagnose the reasoning processes that support it. Figure \ref{fig:overview} summarizes the interactive benchmark and its diagnostic evaluation framework.

Recent work has begun to explore large language models (LLMs) as components of autonomous scientific discovery systems, including closed-loop experimentation, materials design, and scientific hypothesis generation \citep{song2025evaluatinglargelanguagemodels, mitchener2025kosmos, gottweis2025towards, m2024augmenting, boiko2023autonomous, yin2025learning, darvish2025organa}. Some existing agent benchmarks involve iterative interaction with simulated environments \citep{Ciss__2025, song2025evaluatinglargelanguagemodels, macknight2025pre}. However, these benchmarks typically emphasize optimization objectives such as maximizing target material properties or scalar performance metrics rather than hypothesis-driven system discrimination under dynamical feedback. BoxingGym \citep{gandhi2025boxinggym} introduces a benchmark for iterative experimental design across 10 simulated scientific environments, focusing on parameter estimation within generative models. 

Concurrently, \citep{rios2026ai} analyze reasoning traces of 
LLM-based scientific agents using a general vocabulary of six epistemic 
operations across 8 domains, including hypothesis-driven inquiry 
tasks such as inorganic qualitative analysis and spectroscopic structure 
elucidation, which are forms of system identification. They report 
aggregate failure patterns that they argue are largely uniform across 
models.

Despite this progress, no existing benchmark jointly evaluates (i) whether LLM agents can perform hypothesis-driven adaptive experimental planning in a system identification problem, and (ii) which reasoning steps in experiment planning differentiate stronger from weaker agents. \citep{rios2026ai} analyze reasoning traces across broad scientific tasks, spanning workflow execution to hypothesis-driven inquiry, but their focus is not on adaptive experiment planning or on relating planning-specific reasoning steps to per-model task performance. Moreover, they analyze reasoning traces but report aggregate patterns rather than relating specific reasoning capabilities to per-model task performance. 

To address this gap, we introduce a benchmark and diagnostic framework for evaluating hypothesis-driven adaptive experimental planning in a closed-world system identification problem, instantiated using kinetic mechanism identification as a controlled testbed.\footnote{Code: \url{https://github.com/schwallergroup/beyond-accuracy-bench}. Dataset: \url{https://huggingface.co/datasets/xq8wvm/HypothesisDrivenExperimentPlanningBench}.} Our benchmark goes beyond final task accuracy by additionally measuring information gain and exploration behavior. Furthermore, we decompose hypothesis-driven reasoning in a closed-world system identification problem into four steps: belief update, hypothesis discrimination, discriminative experimental design and predictive reasoning under interventions. We systematically evaluate them using an LLM-as-a-judge protocol over reasoning traces, allowing us to diagnose where and how reasoning failures occur.

This paper makes three contributions. (i) We introduce a controlled benchmark for hypothesis-driven adaptive experimental planning in kinetic mechanism identification, where LLM agents iteratively update hypotheses and select new experimental conditions under likelihood-based feedback. (ii) We show that the benchmark differentiates agents by planning performance: the strongest models outperform both adaptive and non-adaptive baselines in task accuracy, and their behavior is not explained by simple space-filling or extreme-condition probing. (iii) We introduce a reasoning-trace diagnostic that decomposes hypothesis-driven experimental planning in closed-world system identification into four steps. This diagnostic reveals that high-performing agents score consistently well across all four steps, whereas weaker agents often succeed at belief update and hypothesis discrimination but score lower on discriminative experiment design and predictive reasoning. Expert annotations support predictive reasoning under our stated criteria, while the evidence for discriminative experiment design is less consistent. These results suggest a hypothesis-operationalization gap: weaker models can recognize plausible hypotheses and articulate their differences, but struggle to convert this understanding into targeted experiments and falsifiable predictions.

\section{Methodology}
\subsection{Benchmark overview}
\label{sec:benchmark_overview}

We formalize adaptive system identification as an interactive agent--environment loop in which
\emph{actions} correspond to experimental conditions and \emph{observations} correspond to noisy time-series trajectories.
The agent's goal is to iteratively design experiments that discriminate among competing candidate mechanisms, thereby reducing uncertainty over the underlying dynamical system. We instantiate this benchmark using kinetic mechanism identification as a concrete testbed.
The benchmark consists of 200 independent reaction systems, each defined by a hidden ground-truth mechanism $m^*$ and associated kinetic parameters $\theta^*$. We consider a hypothesis class of 20 candidate mechanisms $\mathcal{M}=\{m_1,\dots,m_{20}\}$ as described by \citet{bures2023organic}, and construct independent parameterizations per mechanism, yielding 200 total systems. All candidate kinetic mechanisms describe catalytic substrate-to-product transformations, in which a catalyst mediates and accelerates conversion from substrate to product. These mechanisms span common catalytic motifs and their variants, including catalyst dimerization, catalyst activation, catalyst deactivation, and substrate or product inhibition. They therefore define a structured hypothesis class in which mechanisms can share coarse kinetic behavior while differing in specific elementary steps, making discrimination from limited time-series observations nontrivial. Details of the candidate mechanisms are provided in Appendix~\ref{app:candidate_mechanisms}.

For each chemical reaction system, we run an iterative agent–environment loop for kinetic mechanism identification for up to five rounds.
At iteration $k$, the agent proposes an experimental condition $\alpha_k$, corresponding to initial concentrations of substrate, product, and catalyst within predefined ranges.
Given $\alpha_k$, the environment simulates a kinetic experiment by generating a concentration trajectory $\mathbf{C}(t)=(C_S(t),C_P(t))$ under the ground-truth mechanism $m^*$ and parameters $\theta^*$.
From this continuous trajectory, the environment returns a discrete set of noisy time-series observations \(O(\alpha_k)=\{(t_j,\tilde{\mathbf{C}}(t_j;\alpha_k))\}_{j=1}^{6}\), where the six observation times $\{t_j\}$ are sampled at log-uniform intervals and Gaussian measurement noise is added to concentrations. $C_S(t)$ and $C_P(t)$ denote the substrate and product concentrations at time $t$, respectively. \(\tilde{\mathbf{C}}(t_j;\alpha)\) is the measured concentration vector at time \(t_j\).

The environment provides feedback to the agent through negative log-likelihood (NLL). A fitting tool is applied to the accumulated kinetic time-series observations $\mathcal{O}_{1:k} = \{ O(\alpha_1), \dots, O(\alpha_k) \}$. Each candidate mechanism $m_i\in\mathcal{M}$ is fitted independently, producing a set of NLL values
$\{\mathrm{NLL}(m_i)\}_{i=1}^{20}$, which are returned to the agent as structured feedback. 

The agent is implemented as an LLM guided by a fixed system prompt. We intentionally provide only high-level objectives: maintaining plausible hypotheses, selecting informative experiments, and reasoning about how candidate mechanisms would differ under proposed conditions. This allows us to evaluate whether models can autonomously instantiate the necessary reasoning process.
To avoid encoding domain-specific heuristics, no task-specific analysis rules are provided.
The prompt is provided in Appendix~\ref{app:agent_prompt}. 

\subsection{Baselines and tool-based reference}
\label{sec:baselines}

\paragraph{Maximin Latin hypercube sampling non-adaptive baseline}
This baseline provides a non-adaptive space-filling design. For each reaction system, we generate a fixed 5 point Latin hypercube design over 3D condition space and retain the design with the largest minimum pairwise distance in log-transformed space.

\paragraph{Bayesian optimization adaptive baseline} 
After each iteration, we fit a Gaussian process surrogate mapping log-transformed experimental conditions to the resulting mechanism entropy, i.e., the entropy of the candidate-mechanism distribution induced by per-mechanism NLL values (Appendix~\ref{app:metrics_stats}), and select the next experiment by minimizing a lower confidence bound (LCB) acquisition function. The first two experiments are initialized by Latin hypercube sampling.

Maximin Latin hypercube sampling baseline and Bayesian optimization baseline characterize experiment-selection performance based on design-space geometry and NLL feedback based adaptation, respectively. Comparing LLM agents against these baselines allow us to assess whether LLM can translate mechanism information and NLL feedback into effective experiment design by formulating hypotheses about plausible mechanisms and experiments that distinguish among them.

\paragraph{Tool-based frozen-parameter EIG reference}
As an optimistic simulation-tool-based reference, we use a frozen-parameter
approximation to expected information gain. 
We adapt this approximation for computational tractability. Bayesian EIG requires integrating over uncertainty in kinetic parameters and possible future observations for each candidate experiment condition. Our frozen-parameter procedure instead uses fitted point estimates to make model-based experiment selection feasible within the available computational budget. After each iteration, every candidate
mechanism is fitted to the cumulative data, and the resulting best-fit kinetic
constants are reused as fixed representative parameters. For each candidate
condition generated by maximin Latin hypercube sampling, we forward-simulate all
mechanisms with these fixed parameters. Each simulated trajectory is then treated
in turn as a possible future observation, and all mechanisms are scored against
it using the same Gaussian squared-error observation model used for fitting,
without re-fitting kinetic constants to the synthetic trajectory. This yields a
posterior mechanism distribution for each possible source mechanism; the score is
the current mechanism entropy minus the prior-weighted average posterior entropy.
The next experiment is the candidate with the largest score. This is not an exact
Bayesian EIG or a formal upper bound, but serves as an optimistic tool-based
reference because it allows repeated mechanism fitting and forward simulation
during experiment selection. BO therefore tests feedback-only adaptation, whereas this reference uses explicit mechanism models and simulation tools. A natural strengthening would integrate over kinetic-parameter uncertainty and noisy future observations in a posterior-integrated one-step EIG calculation; our CMA-ES fits provide point estimates rather than the parameter posteriors needed for that calculation. The first experiment is initialized by Latin
hypercube sampling.

\subsection{Evaluation metrics}
\label{sec:evaluation_metrics}

\paragraph{Mechanism identification accuracy}
We compute mechanism identification accuracy, defined as the probability that the true mechanism attains the minimum NLL among all candidates when fitted to the accumulated observations.

\paragraph{Mechanism entropy}
We additionally report mechanism entropy over candidates, measuring uncertainty reduction across iterations.

\paragraph{Exploration behavior}
We also report two exploration behavior metrics. \textit{Occupancy entropy} is computed from the cumulative set of proposed conditions after mapping substrate, product, and catalyst concentrations into normalized log space and binning them into a fixed $3 \times 2 \times 3$ grid; we then compute the normalized Shannon entropy of the resulting occupancy distribution.
\textit{S-cat corner mass} is the fraction of cumulative proposals that fall within 15\% of a boundary along all S-cat dimensions in the same normalized log-space representation.
These metrics characterize how each proposer explores the experimental design space, distinguishing diffuse coverage from corner-seeking behavior. Formal definitions of mechanism entropy, occupancy entropy, and S-cat corner mass are provided in Appendix \ref{app:metrics_stats}

\subsection{Reasoning decomposition and analysis}

In closed-world system identification, an agent must use accumulated evidence to distinguish among a fixed set of candidate mechanisms and select informative experiments. We use one natural operational decomposition of this planning process into four interpretable reasoning steps: (i) updating beliefs based on observed evidence (\emph{belief update}), (ii) identifying discriminative differences among competing hypotheses (\emph{hypothesis discrimination}), (iii) selecting experiments that expose these differences (\emph{discriminative experiment design}) and (iv) predicting how these hypotheses respond under interventions (\emph{predictive reasoning}). The agent prompt (Appendix \ref{app:agent_prompt}), which we refer to as the standard prompt, provides only high-level objectives, such as maintaining plausible hypotheses and selecting informative experiments, without prescribing the intermediate reasoning steps required to achieve them.  We therefore analyze whether agents exhibit these intermediate reasoning steps when connecting high-level planning instructions to experimental choices. The steps follow the planning process and are conceptually coupled rather than orthogonal factors. The relationship between prompt instructions and evaluation criteria is summarized in Appendix Table \ref{tab:capabilities}. We additionally evaluate a prompt that explicitly instructs the four reasoning steps. The ablation prompt and results are provided in Appendix \ref{app:reasoning_step_instruction_ablation} and \ref{sec:prompt_effects}.

Each step is assessed using a three-level scoring scheme. Score 0 indicates that the step is not demonstrated; Score 1 indicates partial demonstration; Score 2 indicates full demonstration. While all steps share this three-level structure, the underlying evaluation axis differs across steps, reflecting the distinct nature of each reasoning step. 
While the four-step decomposition is formulated for closed-world system identification, the concrete criteria for each score level are tailored to kinetic mechanism identification.
(i) Belief update is scored on an accuracy axis: whether the candidate set reported by the agent matches the candidate set implied by the NLL feedback. (ii) Hypothesis discrimination is scored on a resolution axis: whether the identified mechanistic differences are at the family level (Score 1) or at the sub-reaction-step level (Score 2). (iii) Discriminative experiment design is scored on an alignment axis: whether the proposed condition is loosely related to the identified mechanistic feature (Score 1) or directly probes it (Score 2). (iv) Predictive reasoning is scored on a resolution axis: whether the predicted experimental outcomes are stated at the family level (Score 1) or as mechanism-specific, observable consequences at the sub-reaction-step level (Score 2). The details of each criterion are shown in Appendix~\ref{app:hypothesis_driven_reasoning_capabilities}.

We score reasoning traces using an LLM-as-a-judge protocol following common practices for rubric-based evaluation \citep{gu2024survey}: a fixed scoring rubric, structured outputs, blinded model identities, and human calibration. Each trace is independently scored on each of the four rubric axes by two LLM judges, GPT-5.4 and Claude Opus 4.7, on a 0/1/2 scale.
To remove direct same-model self-judgments, we exclude the scores assigned by a judge to traces generated by that same model. The final score is the mean of the remaining judge scores. For GPT-5.4 and Claude Opus 4.7 traces, it is the score assigned by the other judge.
For the main standard-prompt analysis, per-model capability scores are computed as the mean across the 120 traces evaluated for that model. The judge receives only the NLL-feedback context from the previous turn, the candidate mechanism definitions, and the model's reasoning response, and returns a discrete score with an evidence-grounded rationale for auditability.

We evaluate 1200 standard-prompt reasoning traces: 120 per model across the 10 evaluated agents, corresponding to 20 ground-truth mechanism families with two kinetic-parameter variants per mechanism, drawn from iterations 2--4 (40 traces per model per iteration). These intermediate rounds require both belief updating from previous feedback and adaptive planning of subsequent experiments. We also apply the LLM judges to the four-step instruction ablation across nine models, comparing 1,078 matched trace pairs from iterations 2--4. Details of the matching and missing traces are provided in Appendix~\ref{app:reasoning_step_instruction_ablation}.

Human validation uses standard-prompt traces only. To verify that the averaged LLM judge tracks human-aligned capability, we initially assess author-LLM agreement on 30 reasoning traces across all four rubric axes; detailed results are provided in Appendix \ref{app:judge_agreement}. We additionally compare LLM-judge scores with annotations from two independent researchers with chemistry expertise on 60 traces: six per model, with three from iteration 3 and three from iteration 4. Annotators were blinded to generating-model identities, LLM-judge scores, and judge rationales. We focus expert validation on discriminative experiment design and predictive reasoning because these axes showed substantial cross-model variation in the initial LLM-judge analysis and correlations with task accuracy (Section \ref{sec:reasoning_analysis}, Appendix \ref{app:judge_agreement}).

\section{Experiments}
\label{sec:experiments}
We evaluate the benchmark using a diverse set of LLM agents, including Gemini 3.1 Pro, Gemini 2.5 Pro, Claude Opus 4.7, Claude 4.5 Haiku, GPT-5.4, GPT-4o, DeepSeek V3.2, Mistral Large 3, Llama 4 Maverick, and Qwen3-32B, with deterministic decoding (temperature = 0). Detailed API configurations and model versions are provided in Appendix~\ref{app:agent_prompt}. Except for the four-step prompting ablation in Section~\ref{sec:prompt_effects}, the LLM-agent results use the standard prompt. Error bars in all plots show 95\% bootstrap confidence intervals for the mean. For each method and metric, we resample the corresponding sample-level values with replacement 4000 times, compute the mean for each bootstrap replicate, and report the 2.5th and 97.5th percentiles. All bootstrap intervals use a fixed random seed of 0. For pooled reasoning-score means, we resample the 40 reaction systems, retaining all three iterations for each selected system. Correlation intervals resample paired model-level aggregates. The prompt-effect tables use paired resampling of shared valid systems; these exploratory intervals are not adjusted for multiple comparisons.

\subsection{Agent adaptive experiment planning vs baselines}

\begin{figure}[h]
    \centering
    \includegraphics[width=0.9\linewidth, trim=0cm 0.0cm 0cm 0.3cm]{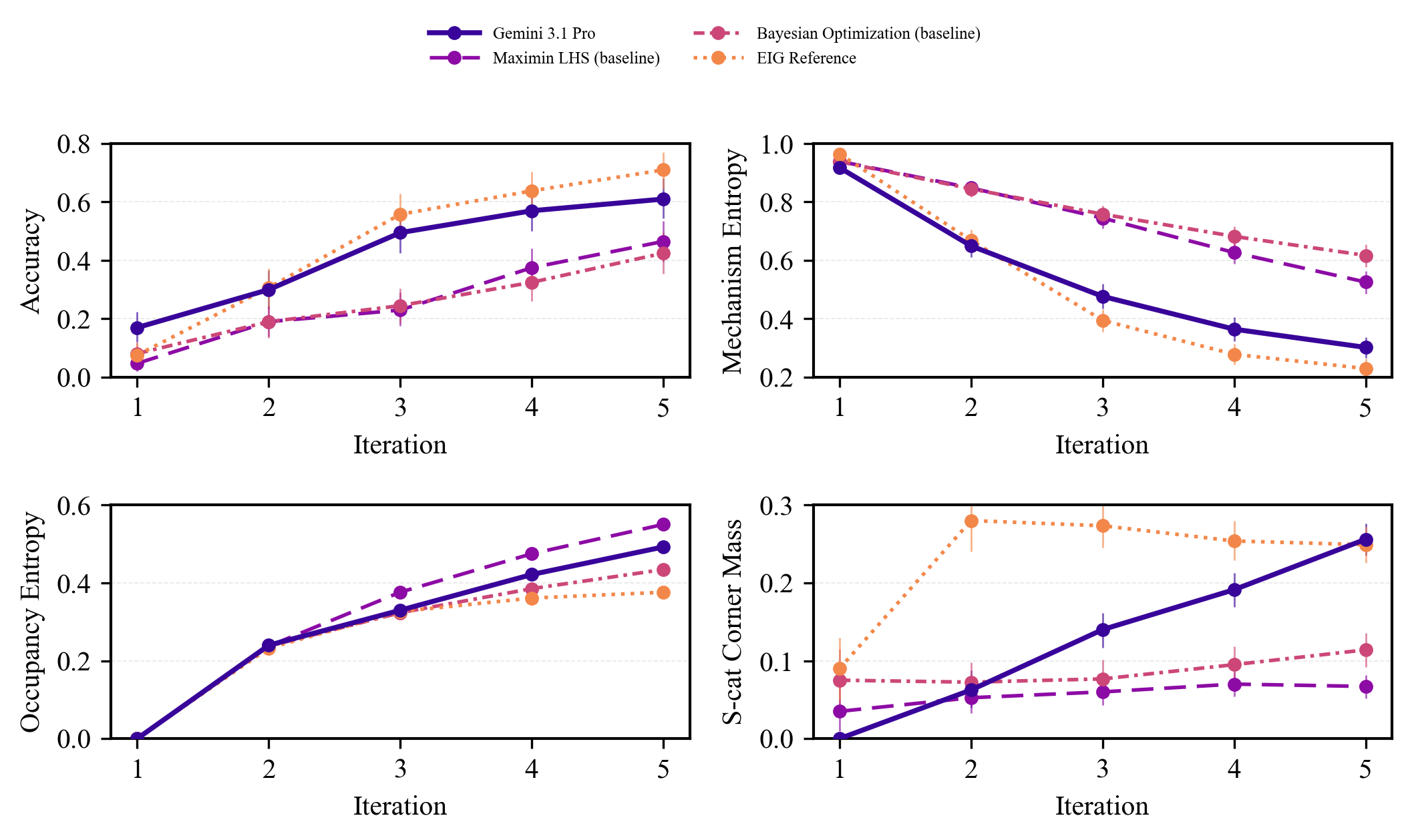}
    \caption{
    \textbf{Top:} Agent versus baselines and tool-based reference. 
    Left: kinetic mechanism identification accuracy over iterations. 
    Right: mechanism entropy over candidate mechanisms. 
    \textbf{Bottom:} Exploration behavior metrics.
    Left: Occupancy entropy over iterations.
    Right: Corner mass over iterations.
    }
    \label{fig:agent_vs_baselines}
\end{figure}

We first analyze the behavior of Gemini 3.1 Pro, the strongest LLM agent in our evaluation (see Section \ref{app:model_comparison}), to characterize how a high-performing agent compares to baselines. Fig. \ref{fig:agent_vs_baselines} shows that Gemini 3.1 Pro achieves higher identification accuracy and lower mechanism entropy than the non-adaptive Maximin LHS baseline and adaptive Bayesian optimization baseline, indicating more effective uncertainty reduction. The tool-based EIG reference, which has access to forward simulation and repeated mechanism fitting during experiment selection, achieves the highest accuracy among the evaluated methods, providing an optimistic tool-assisted reference rather than a formal upper bound.

Gemini 3.1 Pro also shows lower occupancy entropy than Maximin LHS baseline, arguing against the interpretation that its policy is merely broad space-filling exploration. The S-cat corner mass shows that Gemini 3.1 Pro proposes experimental conditions closer to the corners in S-cat space than the Maximin LHS and Bayesian baselines. The EIG reference exhibits a comparable corner mass (0.25), indicating that even an optimistic informative-experiment strategy uses corner regions only for a minority of proposals. Thus, neither the reference nor Gemini 3.1 Pro concentrates all proposals at the corners. Gemini 3.1 Pro's corner mass (0.26) is consistent with informative experiment selection, although these distributional metrics alone do not establish that individual proposals are hypothesis-driven.

Taken together, these diagnostics are consistent with hypothesis-driven adaptive experiment planning: Gemini 3.1 Pro reduces mechanism entropy more effectively than the baselines, while its proposal distribution is neither broad space filling nor degenerate corner seeking.

\subsection{Model comparison}
\label{app:model_comparison}

\begin{figure}[h]
    \centering
    \includegraphics[width=0.7\linewidth, trim=0cm 0.0cm 0cm 0.3cm]{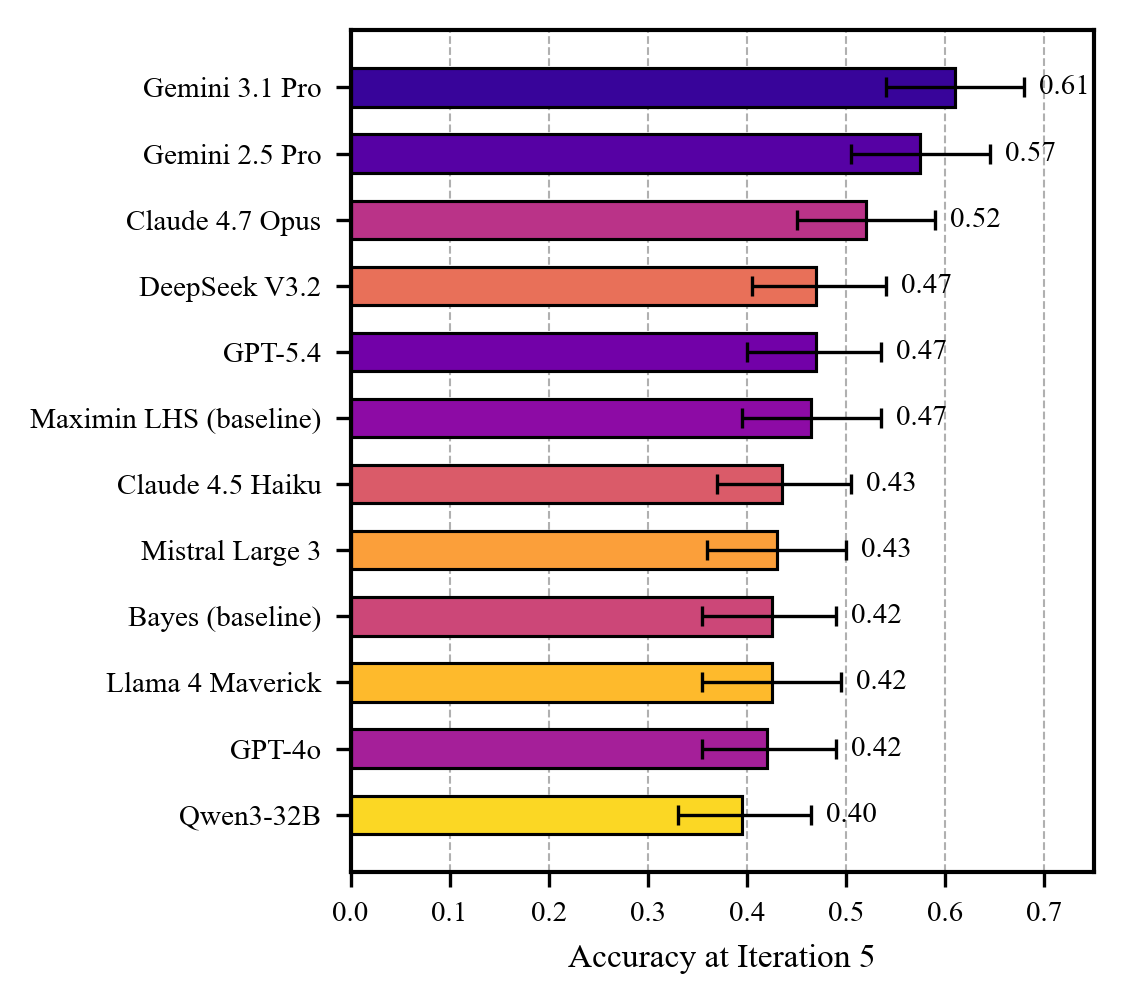}
    \caption{
    Kinetic mechanism identification accuracy at iteration 5 across LLMs and baselines.
    }
    \label{fig:llm_comparison}
\end{figure}

Fig.~\ref{fig:llm_comparison} shows that the benchmark separates models by their performance in adaptive experiment planning capability. At iteration 5, Gemini 3.1 Pro and Gemini 2.5 Pro achieve the highest accuracies, followed by Claude Opus 4.7, all three outperforming both Maximin LHS and Bayesian baselines. GPT-5.4 and DeepSeek V3.2 achieve accuracies comparable to Maximin LHS, the stronger baseline, whereas GPT-4o, Claude 4.5 Haiku, Llama 4 Maverick, and Mistral Large 3 are closer to the Bayesian baseline. Qwen3-32B performs below both baselines. 
These results suggest that the strong models can translate mechanism information and accumulated NLL feedback into more effective experiment selection than Maximin LHS and Bayesian optimization baselines, whereas others do not achieve the same benefit despite having access to this information. To examine where this process succeeds or breaks down, we next analyze reasoning traces through four steps: belief update, hypothesis discrimination, discriminative experiment design, and predictive reasoning.

\subsection{Hypothesis-driven reasoning analysis}
\label{sec:reasoning_analysis}
\begin{figure}[h]
    \centering
    \includegraphics[width=0.9\linewidth, trim=0cm 0.0cm 0cm 0.3cm]{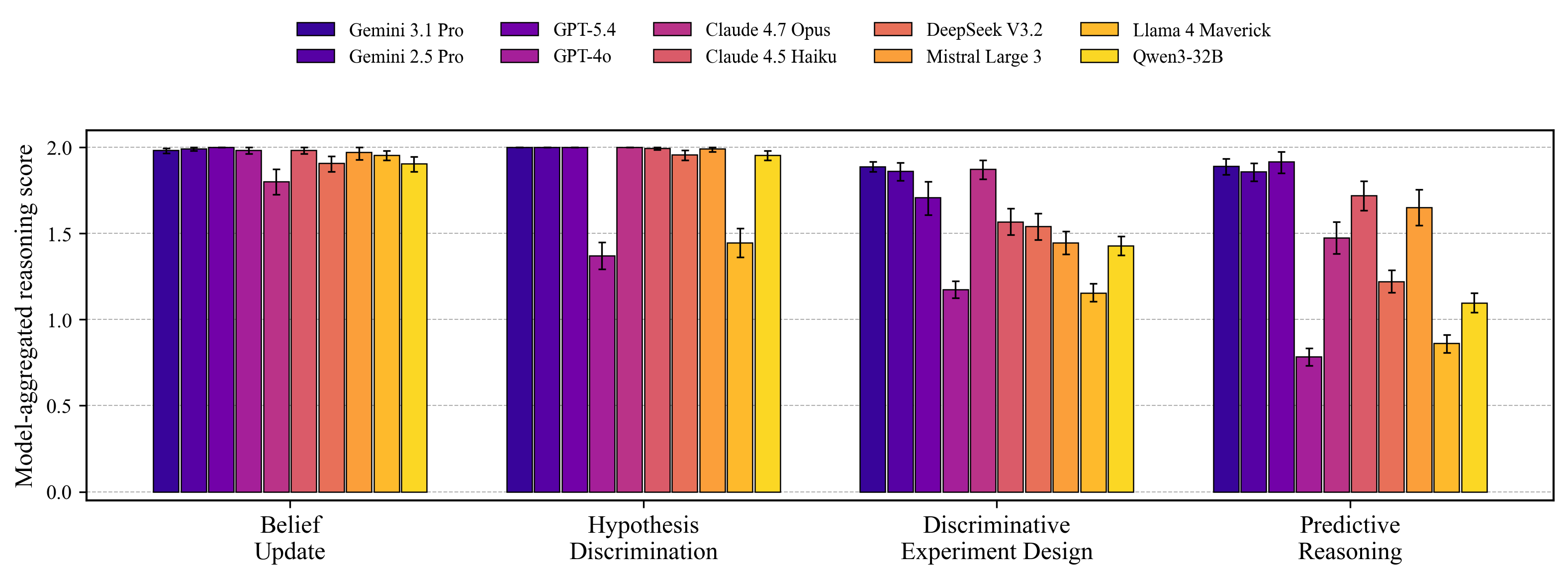}
    
    \vspace{0.3cm}
    
    \includegraphics[width=0.9\linewidth, trim=0cm 0.0cm 0cm 0.3cm]{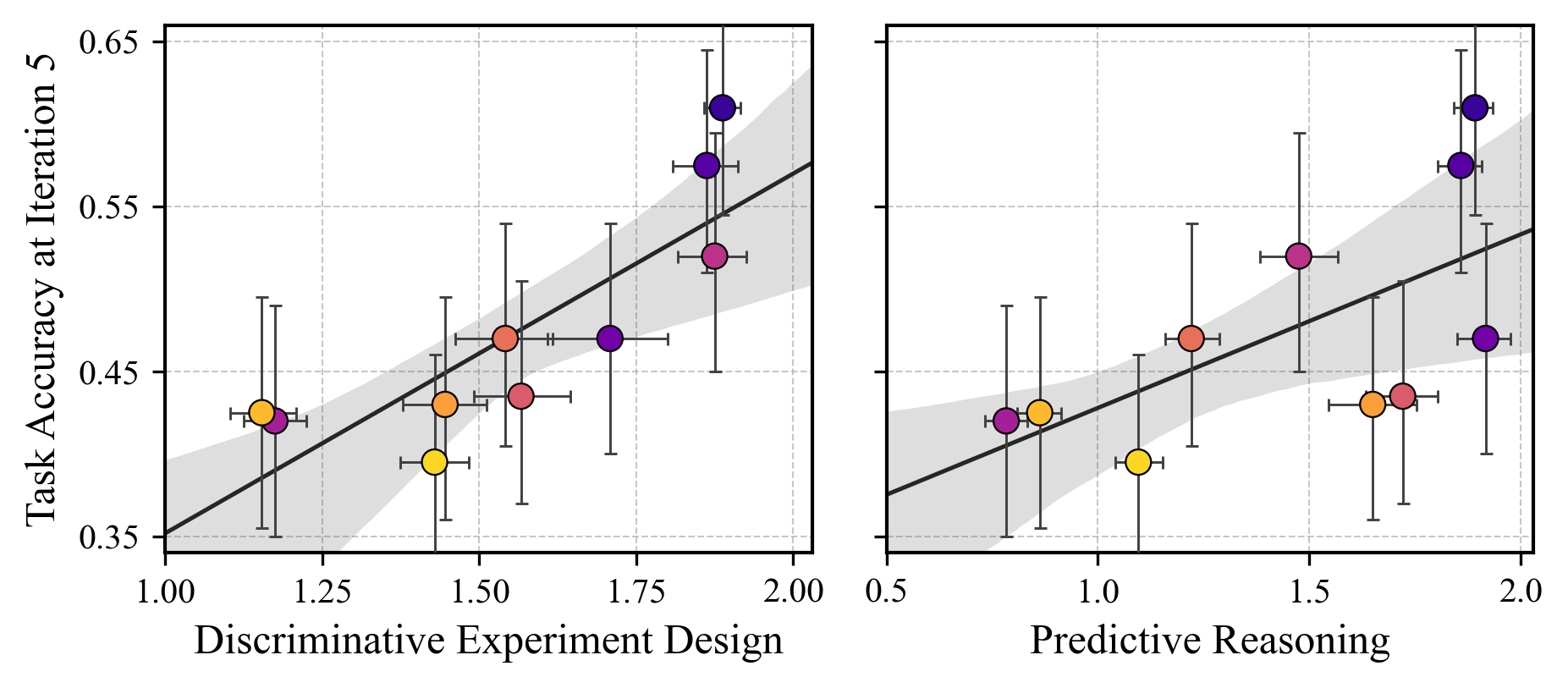}
    \caption{
    \textbf{Top:} Per-model judge scores on the four reasoning steps (belief update, hypothesis discrimination, discriminative experiment design and predictive reasoning) across 10 evaluated LLM agents, using 120 traces per model pooled across iterations 2--4 (40 systems per model, observed at three iterations). Scores exclude same-model self-judgments. Reasoning-score error bars resample reaction systems, keeping their three iterations together.
    \textbf{Bottom:} Per-model judge scores on discriminative experiment design (left) and predictive reasoning (right) versus kinetic mechanism identification accuracy at iteration 5. Thin solid lines show ordinary least-squares fits, with shaded bands indicating 95\% bootstrap confidence intervals. These regression lines are included only as a visual guide.
    }
    \label{fig:reasoning_analysis}
\end{figure}

The top panel of Fig.~\ref{fig:reasoning_analysis} decomposes model reasoning into four steps required for hypothesis-driven experimental planning. Across the four steps, belief update is consistently saturated, hypothesis discrimination is generally high but drops for two lower-accuracy models, GPT-4o and Llama 4 Maverick, and discriminative experiment design and predictive reasoning exhibit substantial cross-model variation. The bottom panel of Fig.~\ref{fig:reasoning_analysis} shows that performance on these latter two capabilities broadly correlates with accuracy. Top-accuracy models score highly on both, while lower-accuracy models tend to score lower. Both discriminative experiment design and predictive reasoning show positive model-level correlations with kinetic mechanism identification accuracy, with Spearman's $\rho=0.92$ for discriminative experiment design (bootstrap 95\% CI: 0.62--1.00) and $\rho=0.74$ for predictive reasoning (bootstrap 95\% CI: 0.22--0.96). This pattern is consistent with a gap between updating or distinguishing hypotheses and translating them into targeted conditions and contrastive predictions. These cross-model associations describe the evaluated agents; they do not establish causal bottlenecks.

Independent expert annotations show positive model-level expert-LLM correlations on the two capabilities with substantial cross-model variance.
For predictive reasoning, expert-LLM agreement is positive for both annotators ($\rho =$0.86, 95\% CI [0.46, 0.99] and 0.92, 95\% CI [0.63, 1.00]). These estimates are close to the expert-expert agreement ($\rho =$ 0.94, 95\% CI [0.76, 1.00]). Predictive reasoning meets our stated validation criteria within this kinetic-mechanism evaluation.
For discriminative experiment design, one expert shows strong agreement with the LLM judge ($\rho =$0.82, 95\% CI [0.54, 0.97]), whereas the other shows moderate agreement and has a wide 95\% CI that includes zero ($\rho =$ 0.58, 95\% CI [-0.12, 0.88]). Thus, the expert-LLM agreement for discriminative experiment design is positive but less consistent than for predictive reasoning.
We interpret the association between LLM-judged discriminative experiment design and task accuracy as suggestive evidence of a reasoning bottleneck.
Trace-level agreement, judge--judge agreement, and self-preference exclusion analyses are reported in Appendix \ref{app:judge_agreement}. The cross-model variation across the four rubric axes is a descriptive finding; interpreting the DED--accuracy association as a bottleneck additionally depends on the diagnostic's validity.

To illustrate what the rubric distinguishes at the trace level, Appendix~\ref{app:reasoning_trace_examples} compares two reasoning traces on the same catalyst-activation sample, whose ground-truth mechanism is M6, that received different scores from the LLM judge. The high-scoring trace received score 2 on all four capabilities. The lower-scoring trace received the score 2 for belief update and hypothesis discrimination, but score 1 for discriminative experiment planning and predictive reasoning. Both traces correctly identified the same hypothesis space: catalyst-activation mechanisms (M6/M8), a substrate-assisted mechanism (M7), and a catalyst-dimer mechanism (M3). The high-scoring trace converted this hypothesis space into a diagnostic intervention: it proposed a substrate-starvation experiment to separate M7, which requires an additional substrate-binding step to form catS2, from M6/M8, which rely on activated catalyst cat*. It also made concrete predictions, namely that M7 should slow or stall under substrate starvation whereas M6/M8 should retain activity through catalyst activation.

In contrast, the lower-scoring trace updated the hypothesis space correctly and identified the M6--M8 distinction as irreversible versus reversible catalyst activation, but its proposed intervention was less diagnostic. It added initial product to test whether the system could ``reset'' under M8, even though product addition is not directly tied to the main activation-reversibility contrast between M6 and M8. Its predictions also remained underspecified: it stated that product might affect M8 but not M6 and that catalyst changes might amplify activation effects, without specifying the kinetic signature expected for each mechanism. This example illustrates what the planning and prediction axes capture: low scores are not assigned because the model names the wrong mechanisms, but because it fails to translate the mechanism features into a targeted experiment and falsifiable, mechanism-specific predictions.

\subsection{Effect of explicit reasoning instructions}
\label{sec:prompt_effects}

\begin{figure*}[h]
    \centering
    \includegraphics[width=\textwidth]{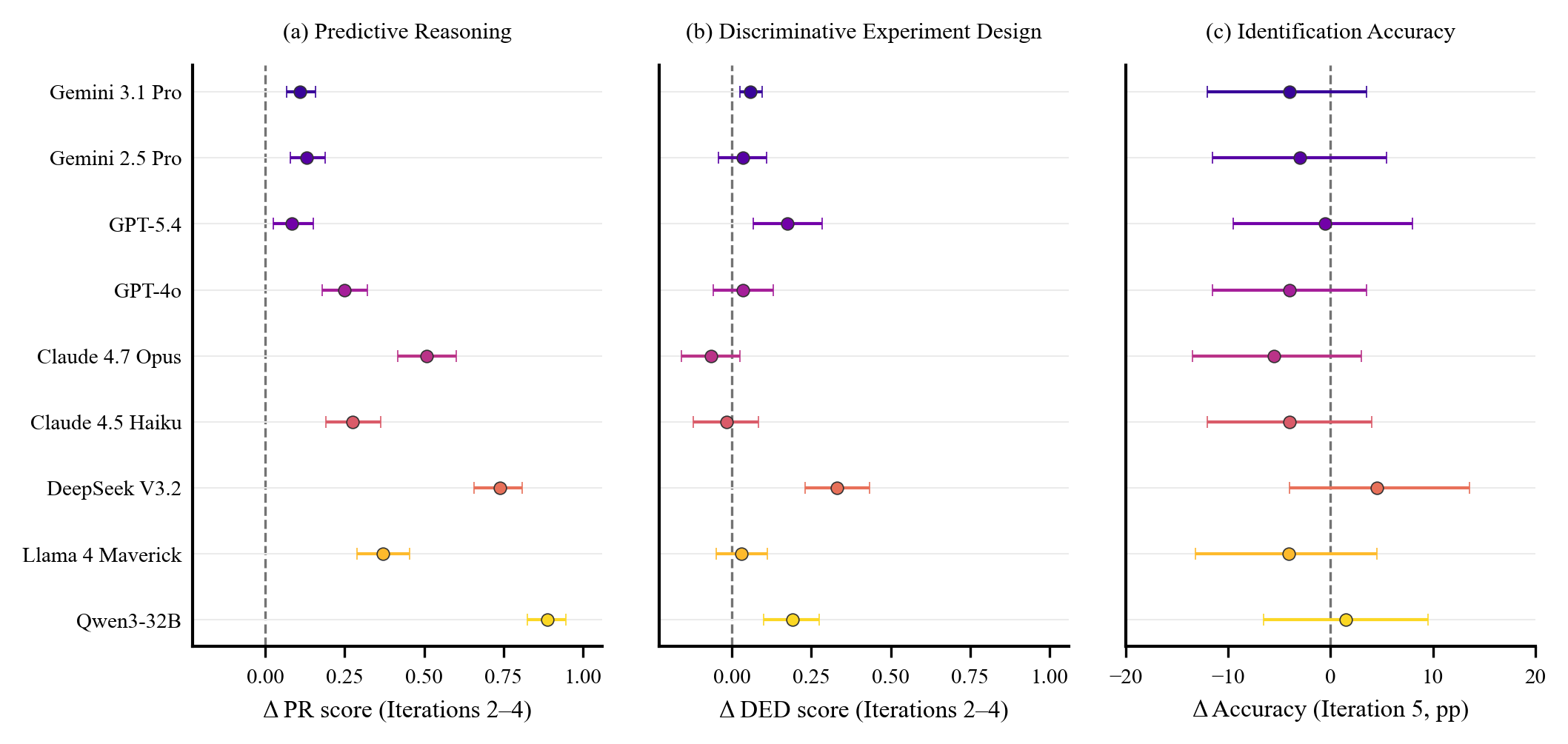}
    \caption{
        Paired effects of explicit four-step prompting across nine models.
        Panels show changes in (a) predictive reasoning (PR) scores and
        (b) discriminative experiment design (DED) scores pooled across iterations~2--4,
        and (c) identification accuracy at iteration~5.
        All changes are computed as four-step minus standard prompting.
        Points indicate mean paired differences; error bars show 95\% percentile bootstrap confidence intervals.
    }
    \label{fig:paired_effects_four_step_prompting}
    \label{fig:prompting-ablation-paired-changes}
\end{figure*}

To test whether the observed reasoning weaknesses can be mitigated by prompting, we evaluated explicit four-step instructions across nine models (Appendix~\ref{app:reasoning_step_instruction_ablation}). 
Figure \ref{fig:paired_effects_four_step_prompting} shows paired effects of explicit four-step prompting across nine models. Predictive-reasoning point estimates increase for all nine models, whereas changes in discriminative experiment design are smaller and heterogeneous. These changes in judged reasoning do not establish better task performance: paired accuracy changes range from $-5.5$ to $+4.5$ percentage points, and all nine 95\% confidence intervals include zero (Table~\ref{tab:step_prompt_accuracy}). The intervals establish neither an accuracy improvement nor equivalence between prompts. Within the four-step condition, discriminative experiment design remains positively associated with accuracy ($\rho=0.971$, 95\% CI [0.762, 1.000]), while the predictive reasoning association is more uncertain ($\rho=0.591$, 95\% CI [-0.193, 0.930]).

This ablation illustrates why diagnostic scores and task outcomes should be evaluated together. More explicit contrastive predictions can improve a reasoning trace's rubric score without demonstrating that the proposed intervention better separates the candidate mechanisms. We therefore report paired changes in identification accuracy and proposal-space behavior alongside the reasoning scores, including valid-system counts and missing proposals. The diagnostic identifies behaviors to inspect and test; a prompt intervention must also be evaluated against the downstream task objective.

\section{Conclusion}
\label{sec:conclusion}

In this work, we introduced a benchmark and diagnostic evaluation framework for hypothesis-driven experimental planning in a controlled kinetic-mechanism identification setting. Rather than evaluating agents only by final identification accuracy, our framework analyzes both task-level behavior and the intermediate reasoning processes that support experimental planning. Stronger agents outperform non-adaptive and adaptive baselines in final identification accuracy, and their proposal distributions are not fully explained by simple space-filling or corner-seeking heuristics, suggesting behavior consistent with hypothesis-driven adaptive experiment design. At the reasoning-process level, belief update and hypothesis discrimination are largely saturated across models, whereas discriminative experiment design and predictive reasoning exhibit substantial variation and broadly track task performance. 
Predictive-reasoning scores meet our stated validation criteria in this setting; expert agreement for discriminative experiment design is positive but less consistent, so its association with accuracy remains suggestive.
Taken together, these findings suggest that weaker models often recognize plausible hypotheses and mechanistic differences, but struggle to translate them into contrastive predictions about experimental outcomes, with more tentative evidence of difficulty translating those differences into actionable experimental design. 
These findings highlight the need to evaluate not only outcomes, but also the reasoning processes that lead to them. Moreover, analyzing reasoning steps at the per-model level enables identifying which specific reasoning steps differentiate stronger from weaker agents and which may serve as bottlenecks in this setting. We hope this work motivates structured evaluation of reasoning processes in scientific domains, supporting both model development and practical deployment of LLM agents for scientific experimentation.

\section{Limitations}
\label{sec:limitation}
Our evaluation framework diagnoses hypothesis-driven adaptive experimental planning in closed-world system identification, where candidate mechanisms are fixed and known. It does not evaluate open-world hypothesis generation or revision of the hypothesis space. All empirical evidence comes from kinetic mechanism identification; the rubric criteria are domain-specific, and cross-domain transfer and behavior beyond the five-round budget remain untested.

The judge scores the reasoning trace without observing the realized informativeness of the proposed experiment. Plausible articulation can therefore receive a high score without guaranteeing an informative intervention. Explicit four-step prompting improves predictive-reasoning scores but does not establish that higher rubric scores cause better experiment selection. Independent validation uses 60 traces and ten model aggregates: predictive reasoning meets our stated criteria in this setting, whereas discriminative experiment design agreement is mixed and its low-score endpoint remains insufficiently assessed. Exclusion analyses address direct self-judging, but no third disjoint judge was evaluated.

The synthetic benchmark is intended for controlled evaluation rather than direct assessment of wet-lab readiness. Its mechanism templates, parameter ranges, simulator assumptions, and filtering of sampled systems limit coverage. The hosted dataset's Croissant metadata documents these limitations, potential biases, intended uses, and social-impact considerations. Future work should test broader domains, longer budgets, and posterior-integrated experimental-design references.

\begin{ack}
Philippe Schwaller and Amin Mansouri acknowledge support from the Swiss National Science Foundation (SNSF Starting Grant No.~226509). The authors also sincerely thank Xuan-Vu Nguyen and Saumya Thakur for their valuable assistance with annotating the reasoning traces. The authors declare no competing interests.
\end{ack}

\newpage
\bibliographystyle{plainnat}
\bibliography{references}

\newpage
\appendix
\section*{Appendix} 

\section{Candidate mechanisms}
\label{app:candidate_mechanisms}
\subsection{Notation and modeling assumptions}
We model all mechanisms as sets of elementary reactions and derive dynamics under standard mass-action kinetics. Reversibility is specified solely by the arrow type: \texttt{<=>} denotes a reversible elementary step and \texttt{=>} denotes an irreversible elementary step. Observations include only substrate and product concentrations ($C_S(t)$ and $C_P(t)$); all catalyst-related intermediates are latent. Initial conditions: $C_S(0)$, $C_P(0)$, and $C_{\mathrm{cat}}(0)$ are chosen by the agent/baseline. All intermediate complexes (e.g., \texttt{catS}, \texttt{catP}, \texttt{cat2}, \texttt{cat*}, and all \texttt{inactive\_*} species) are initialized to $0$. Additional species that appear only in certain mechanisms are initialized to $1$: $C_X(0)=1$, $C_{\mathrm{inhibitor}}(0)=1$, and $C_L(0)=1$.

\subsection{Candidate Mechanisms}
These mechanisms are grouped into four broad classes: (a) a core mechanism without side reactions, (b) mechanisms containing bicatalytic steps, (c) mechanisms involving catalyst activation, and (d) mechanisms with catalyst deactivation pathways.
\paragraph{Category mapping}
We use the following mapping between candidate mechanisms and the four mechanism classes: (a) \texttt{M1}; (b) \texttt{M2-M5}; (c) \texttt{M6-M8}; and (d) \texttt{M9-M20}.

\paragraph{Mechanism definitions}
Each mechanism is specified by a sequence of elementary reactions in a compact reaction-string form. The ODEs are derived under mass-action kinetics from the reaction graph and arrow directions alone (reversible \texttt{<=>} vs irreversible \texttt{=>}). We use \texttt{|} to separate elementary steps within a mechanism.
\begin{PromptBox}
M1:  S + cat <=> catS | catS <=> P + cat
M2:  S + cat <=> catS | catS <=> P + cat | 2cat <=> cat2
M3:  S + cat2 <=> cat2S | cat2S <=> P + cat2 | 2cat <=> cat2
M4:  X + catS <=> S + cat | X + catS <=> P + cat
M5:  S + cat <=> catS | catS + cat <=> catP + cat | catP <=> P + cat
M6:  cat => cat* | S + cat* <=> cat*S | cat*S <=> P + cat*
M7:  S + cat <=> catS | S + catS <=> catS2 | catS2 <=> P + catS
M8:  S + cat* <=> cat*S | cat*S <=> P + cat* | cat <=> cat* + L
M9:  S + cat <=> catS | catS <=> P + cat | cat => inactive_cat
M10: S + cat <=> catS | catS <=> P + cat | inhibitor + cat => inactive_catI
M11: S + cat <=> catS | catS <=> P + cat | S + cat => inactive_catS
M12: S + cat <=> catS | catS <=> P + cat | P + cat => inactive_catP
M13: S + cat <=> catS | catS <=> P + cat | 2cat => inactive_cat2
M14: S + cat <=> catS | catS <=> P + cat | catS => inactive_catS
M15: S + cat <=> catS | catS <=> P + cat | inhibitor + catS => inactive_catSI
M16: S + cat <=> catS | catS <=> P + cat | S + catS => inactive_catS2
M17: S + cat <=> catS | catS <=> P + cat | P + catS => inactive_catSP
M18: S + cat <=> catS | catS <=> P + cat | 2catS => inactive_cat2S2
M19: S + cat <=> catS | catS <=> P + cat | cat + catS => inactive_cat2S
M20: S + cat <=> catS | catS <=> P + cat | cat => inactive_cat | catS => inactive_catS
\end{PromptBox}

\paragraph{Mechanism definition JSON provided to agents}
\label{app:mechanism_definition_json}
\begin{PromptBox}
{
  "M1": "S+cat<=>catS|catS<=>P+cat",
  "M2": "S+cat<=>catS|catS<=>P+cat|2cat<=>cat2",
  "M3": "S+cat2<=>cat2S|cat2S<=>P+cat2|2cat<=>cat2",
  "M4": "X+catS<=>S+cat|X+catS<=>P+cat",
  "M5": "S+cat<=>catS|catS+cat<=>catP+cat|catP<=>P+cat",
  "M6": "cat=>cat*|S+cat*<=>cat*S|cat*S<=>P+cat*",
  "M7": "S+cat<=>catS|S+catS<=>catS2|catS2<=>P+catS",
  "M8": "S+cat*<=>cat*S|cat*S<=>P+cat*|cat<=>cat*+L",
  "M9": "S+cat<=>catS|catS<=>P+cat|cat=>inactive_cat",
  "M10": "S+cat<=>catS|catS<=>P+cat|inhibitor+cat=>inactive_catI",
  "M11": "S+cat<=>catS|catS<=>P+cat|S+cat=>inactive_catS",
  "M12": "S+cat<=>catS|catS<=>P+cat|P+cat=>inactive_catP",
  "M13": "S+cat<=>catS|catS<=>P+cat|2cat=>inactive_cat2",
  "M14": "S+cat<=>catS|catS<=>P+cat|catS=>inactive_catS",
  "M15": "S+cat<=>catS|catS<=>P+cat|inhibitor+catS=>inactive_catSI",
  "M16": "S+cat<=>catS|catS<=>P+cat|S+catS=>inactive_catS2",
  "M17": "S+cat<=>catS|catS<=>P+cat|P+catS=>inactive_catSP",
  "M18": "S+cat<=>catS|catS<=>P+cat|2catS=>inactive_cat2S2",
  "M19": "S+cat<=>catS|catS<=>P+cat|cat+catS=>inactive_cat2S",
  "M20": "S+cat<=>catS|catS<=>P+cat|cat=>inactive_cat|catS=>inactive_catS"
}
\end{PromptBox}

In addition to \texttt{S}, \texttt{P}, and \texttt{cat}, the candidate mechanism set introduces the following extra species that are present at initialization: \texttt{X} (a different catalytic species, used in \texttt{M4}), \texttt{L} (a ligand, used in \texttt{M8}), and \texttt{inhibitor} (an inhibitor species, used in \texttt{M10} and \texttt{M15}). All other species in the reaction strings are mechanism-specific intermediates (including activated/deactivated forms and complexes) that are initialized to $0$ and are defined implicitly by the reaction topology above.

\subsection{Parameter priors and constraints}
All kinetic constants are sampled independently from a log-uniform prior, $\theta \sim \mathrm{LogUniform}(10^{-2}, 10^{2})$. After sampling, we filter parameter sets using mechanism-specific dynamical conditions to ensure that the defining catalyst-related species for each mechanism attain non-negligible concentrations under a reference simulation.

We define conversion as fractional substrate consumption, $x(t)=\big(C_S(0)-C_S(t)\big)/C_S(0)$. For mechanisms with multiple catalyst-containing species, we define the total catalyst pool as $C_{\mathrm{tot}}(t)=\sum_{s\in\mathcal{S}_{\mathrm{cat}}} C_s(t)$, where $\mathcal{S}_{\mathrm{cat}}$ is the set of catalyst-related species for that mechanism, and define a catalyst-species fraction for any numerator set $\mathcal{N}$ as
\[
r_{\mathcal{N}}(t) \;=\; \frac{\sum_{s\in\mathcal{N}} C_s(t)}{C_{\mathrm{tot}}(t)} \, .
\]

\paragraph{Reference simulation conditions for filtering}

For each sampled parameter set, we evaluate filtering on 100 validation
initial conditions sampled from the experimental design space using Latin
hypercube sampling. The unit-cube samples are mapped to concentration values
using log-uniform scaling:
\[
C_S(0) \sim \mathrm{LogUniform}(0.1, 5.0), \qquad
C_{\mathrm{cat}}(0) \sim \mathrm{LogUniform}(5\times10^{-4}, 0.2).
\]
For product, whose lower bound is zero, we use an epsilon-shifted log-uniform
mapping over
\[
C_P(0) \in [0, 0.5],
\]
with $\epsilon=10^{-9}$.
A parameter set is accepted if at least 5\% of these validation conditions pass
both the product-formation filter and the mechanism-specific dynamical
conditions as follows.

\paragraph{Product-formation filter}

For a validation condition with initial concentrations $C_S(0)$ and $C_P(0)$,
we require
\[
0.5\{C_S(0)+C_P(0)\} \le C_P(t_{\mathrm{eff}}) \le C_S(0)+C_P(0),
\]
where $t_{\mathrm{eff}}$ is the first time at which $C_P(t)$ reaches 95\% of
its value at the end of the dense reference simulation.

\paragraph{Mechanism-specific filters}

The mechanism-specific filtering criteria below are adopted from \cite{bures2023organic}.

\begin{itemize}
    \item \texttt{M1}, \texttt{M4}, \texttt{M5}: no additional filtering beyond the prior.

    \item \texttt{M2} ($\mathcal{S}_{\mathrm{cat}}=\{\texttt{cat},\texttt{catS},\texttt{cat2}\}$): in a reference run with 5 mol\% catalyst, require a significant dimeric fraction during the early-to-mid reaction, i.e., $\max_{x\in[0.2,0.5]} r_{\{\texttt{cat2}\}} \ge 0.1$.

    \item \texttt{M3} ($\mathcal{S}_{\mathrm{cat}}=\{\texttt{cat},\texttt{cat2},\texttt{cat2S}\}$): in a reference run with 5 mol\% catalyst, require that free \texttt{cat} is not vanishingly small at the start of the reaction, i.e., $\min_{x\in[0.0,0.1]} r_{\{\texttt{cat}\}} \ge 0.05$.

    \item \texttt{M6} ($\mathcal{S}_{\mathrm{cat}}=\{\texttt{cat},\texttt{cat*},\texttt{cat*S}\}$): for at least one catalyst loading in 3--7 mol\%, require that activated/complexed catalyst has an intermediate fraction at 20\% conversion, i.e., $r_{\{\texttt{cat*},\texttt{cat*S}\}} \in [0.1,0.9]$ at $x=0.2$.

    \item \texttt{M7} ($\mathcal{S}_{\mathrm{cat}}=\{\texttt{cat},\texttt{catS},\texttt{catS2}\}$): for at least one catalyst loading in 3--7 mol\%, require $r_{\{\texttt{catS},\texttt{catS2}\}} \in [0.1,0.8]$ at $x=0.2$.

    \item \texttt{M8} ($\mathcal{S}_{\mathrm{cat}}=\{\texttt{cat},\texttt{cat*},\texttt{cat*S}\}$): for at least one catalyst loading in 3--7 mol\%, require $r_{\{\texttt{cat*},\texttt{cat*S}\}} \in [0.1,0.9]$ at $x=0.5$.

    \item \texttt{M9} ($\mathcal{S}_{\mathrm{cat}}=\{\texttt{cat},\texttt{catS},\texttt{inactive\_cat}\}$): for at least one catalyst loading in 3--7 mol\%, require an intermediate active-catalyst fraction, $r_{\{\texttt{cat},\texttt{catS}\}} \in [0.5,0.9]$ at $x=0.5$.

    \item \texttt{M10} ($\mathcal{S}_{\mathrm{cat}}=\{\texttt{cat},\texttt{catS},\texttt{inactive\_catI}\}$): for at least one catalyst loading in 3--7 mol\%, require appreciable inhibited catalyst, $r_{\{\texttt{inactive\_catI}\}} \ge 0.1$ at $x=0.5$.

    \item \texttt{M11} ($\mathcal{S}_{\mathrm{cat}}=\{\texttt{cat},\texttt{catS},\texttt{inactive\_catS}\}$): for at least one catalyst loading in 3--7 mol\%, require $r_{\{\texttt{cat},\texttt{catS}\}} \in [0.5,0.9]$ at $x=0.5$.

    \item \texttt{M12} ($\mathcal{S}_{\mathrm{cat}}=\{\texttt{cat},\texttt{catS},\texttt{inactive\_catP}\}$): for at least one catalyst loading in 3--7 mol\%, require $r_{\{\texttt{cat},\texttt{catS}\}} \in [0.5,0.9]$ at $x=0.5$.

    \item \texttt{M13} ($\mathcal{S}_{\mathrm{cat}}=\{\texttt{cat},\texttt{catS},\texttt{inactive\_cat2}\}$): for at least one catalyst loading in 3--7 mol\%, require $r_{\{\texttt{cat},\texttt{catS}\}} \in [0.5,0.8]$ at $x=0.5$.

    \item \texttt{M14} ($\mathcal{S}_{\mathrm{cat}}=\{\texttt{cat},\texttt{catS},\texttt{inactive\_catS}\}$): for at least one catalyst loading in 3--7 mol\%, require $r_{\{\texttt{cat},\texttt{catS}\}} \in [0.5,0.9]$ at $x=0.5$.

    \item \texttt{M15} ($\mathcal{S}_{\mathrm{cat}}=\{\texttt{cat},\texttt{catS},\texttt{inactive\_catSI}\}$): for at least one catalyst loading in 3--7 mol\%, require $r_{\{\texttt{inactive\_catSI}\}} \ge 0.1$ at $x=0.5$.

    \item \texttt{M16} ($\mathcal{S}_{\mathrm{cat}}=\{\texttt{cat},\texttt{catS},\texttt{inactive\_catS2}\}$): for at least one catalyst loading in 3--7 mol\%, require $r_{\{\texttt{cat},\texttt{catS}\}} \in [0.5,0.9]$ at $x=0.5$.

    \item \texttt{M17} ($\mathcal{S}_{\mathrm{cat}}=\{\texttt{cat},\texttt{catS},\texttt{inactive\_catSP}\}$): for at least one catalyst loading in 3--7 mol\%, require $r_{\{\texttt{cat},\texttt{catS}\}} \in [0.5,0.9]$ at $x=0.5$.

    \item \texttt{M18} ($\mathcal{S}_{\mathrm{cat}}=\{\texttt{cat},\texttt{catS},\texttt{inactive\_cat2S2}\}$): for at least one catalyst loading in 3--7 mol\%, require $r_{\{\texttt{cat},\texttt{catS}\}} \in [0.5,0.8]$ at $x=0.5$.

    \item \texttt{M19} ($\mathcal{S}_{\mathrm{cat}}=\{\texttt{cat},\texttt{catS},\texttt{inactive\_cat2S}\}$): for at least one catalyst loading in 3--7 mol\%, require $r_{\{\texttt{cat},\texttt{catS}\}} \in [0.5,0.8]$ at $x=0.5$.

    \item \texttt{M20} ($\mathcal{S}_{\mathrm{cat}}=\{\texttt{cat},\texttt{catS},\texttt{inactive\_cat},\texttt{inactive\_catS}\}$): require (i) an intermediate active-catalyst fraction, $r_{\{\texttt{cat},\texttt{catS}\}} \in [0.5,0.9]$ at $x=0.5$, and (ii) appreciable deactivated fractions, $r_{\{\texttt{inactive\_cat}\}} > 0.05$ and $r_{\{\texttt{inactive\_catS}\}} > 0.05$ at $x=0.5$.
\end{itemize}

\section{Simulator and data generation}
\label{app:simulator_data_generation}
\subsection{Experimental condition space}
The experimental condition is $\alpha = \big(C_S(0),\, C_{\mathrm{cat}}(0),\, C_P(0)\big)$, i.e., the initial concentrations of substrate, catalyst, and product. Allowed ranges are the same for the agent and the random baseline: $C_S(0)\in[0.1, 5.0]$, $C_{\mathrm{cat}}(0)\in[5\times 10^{-4}, 0.2]$, and $C_P(0)\in[0, 0.5]$. The non-adaptive random baseline samples conditions from this domain using log-uniform sampling: $C_S(0)$ and $C_{\mathrm{cat}}(0)$ are sampled log-uniformly over their strictly-positive ranges; for $C_P(0)$, we sample log-uniformly over $[\varepsilon, 0.5+\varepsilon]$ with $\varepsilon=10^{-9}$ and then set $C_P(0)\leftarrow C_P(0)-\varepsilon$ to map samples back to $[0,0.5]$.

\subsection{ODE simulation and observation times}
We first simulate a dense trajectory up to a fixed maximum time \texttt{max\_t}. Let $P_{\infty}=P(\texttt{max\_t})$ denote the final product concentration at the end of this dense simulation, and define the effective horizon as the earliest time $t_{\max}\in[0,\texttt{max\_t}]$ such that $P(t_{\max})\ge 0.95\,P_{\infty}$. We then sample 6 sparse observation points log-uniformly on $[t_{\mathrm{start}}, t_{\max}]$ 
with $t_{\mathrm{start}}=10^{-6}\,t_{\max}$ (and using machine epsilon if $t_{\mathrm{start}}=0$). Kinetic time-series data are generated using COPASI \citep{hoops2006copasi} via \texttt{basico.run\_time\_course} with \texttt{num\_steps=5000}, \texttt{output\_event=True}, \texttt{max\_steps=1,000,000}, absolute tolerance $10^{-12}$, and relative tolerance $10^{-8}$; solver exceptions are treated as failed simulations. When computing NLL values by fitting candidate mechanisms, we integrate ODEs with SciPy \citep{2020SciPy-NMeth} \texttt{solve\_ivp} using \texttt{method=LSODA} with \texttt{rtol=1e-4} and \texttt{atol=1e-8} at the experimental time points.

\subsection{Observation model}
Noisy observations are obtained by adding Gaussian noise to the concentrations at each observation time. The observation noise level $\sigma=0.05$ is defined with respect to the normalized concentration scale used in the simulation and is applied consistently across all experiments. Specifically, the observed concentrations are given by
\[
\tilde{\mathbf{C}}(\tau_j;\alpha) = \mathbf{C}(\tau_j;\alpha) + \boldsymbol{\varepsilon}_j,
\quad
\boldsymbol{\varepsilon}_j \sim \mathcal{N}(\mathbf{0}, \sigma^2 \mathbf{I}).
\]

\section{Likelihood-Based Evaluation and Parameter Fitting}
\label{app:fitting}

\subsection{Gaussian Observation Model and NLL}
We define a Gaussian observation model with independent noise across data points. Let $\mathrm{SSE}$ denote the summed squared error between simulated and observed concentrations across the accumulated collection of kinetic time-series observations (i.e., across all iterations included in $\mathcal{O}_{1:k}$, all observed species, and all observation time points), and let $n$ be the total number of scalar data points. The log-likelihood used in our implementation is
\[
\log p(\mathcal{O}_{1:k}\mid \theta, \sigma)
= -\tfrac{1}{2} n \log (2\pi \sigma^2) - \tfrac{1}{2}\,\mathrm{SSE}/\sigma^2.
\]
We report negative log-likelihood (NLL) as $\mathrm{NLL}= -\log p(\mathcal{O}_{1:k}\mid \theta, \sigma)$.

During fitting, $\sigma$ is treated as a free parameter and is optimized jointly with kinetic constants. We parameterize $\theta$ and $\sigma$ in log-space with uniform box constraints: $\log_{10} \theta \in [-2, 2]$ and $\log_{10}\sigma \in [-2.5, -1.5]$.

\subsection{Parameter estimation via CMA-ES}
Parameter estimation is performed using the Covariance Matrix Adaptation Evolution Strategy (CMA-ES) \cite{hansen2016cma}, which is well suited for nonlinear, nonconvex optimization problems arising in kinetic modeling. All fitting procedures use identical optimization settings across mechanisms.
The objective minimized by CMA-ES is the NLL defined above. We use a log-parameterization with box constraints for $\log_{10}\theta$ and $\log_{10}\sigma$ as specified above and initialize CMA-ES with $\sigma_0=1.0$ in the internal parameter space. CMA-ES options used in our implementation are:
\begin{itemize}
    \item \texttt{ftarget=-10000} (a sufficiently small target so optimization effectively runs as unconstrained minimization),
    \item \texttt{tolstagnation=20},
    \item \texttt{popsize\_factor=7},
    \item \texttt{timeout=60} seconds,
    \item \texttt{verb\_disp=0}.
\end{itemize}
Optimizer failures (including ODE integration failures during objective evaluation) are treated as fitting failures.

\section{Agents}
\subsection{Interaction protocol}
After each proposed experiment, the environment provides the proposed initial conditions and a JSON object of per-mechanism NLL values for the 20 candidates indexed by \texttt{"M1"}--\texttt{"M20"} (lower NLL indicates a better fit to the accumulated observations); if fitting fails for all mechanisms for a given condition, NLL values are returned as \texttt{null} for that iteration (see Appendix~\ref{app:failures_and_edge_cases}). The agent proposes the next experimental condition in a \texttt{<suggestion>...</suggestion>} JSON object with keys \texttt{"initial\_concentration\_of\_substrate"}, \texttt{"initial\_concentration\_of\_catalyst"}, and \texttt{"initial\_concentration\_of\_product"} (within the allowed ranges), and outputs a \texttt{<candidate\_mechanisms>} field containing mechanism names (\texttt{M1}--\texttt{M20}) for diagnostics.

\subsection{Prompts and LLM configuration}
\label{app:agent_prompt}
We run agents via OpenRouter (\texttt{https://openrouter.ai/api/v1}) using the following model identifiers: \texttt{google/gemini-3.1-pro}, \texttt{google/gemini-2.5-pro}, \texttt{anthropic/claude-opus-4.7}, \texttt{anthropic/claude-haiku-4.5}, \texttt{openai/gpt-5.4}, \texttt{openai/gpt-4o}, \texttt{deepseek/deepseek-r1}, \texttt{mistralai/mistral-large-2512}, \texttt{meta-llama/llama-4-maverick},
and \texttt{qwen/qwen3-32b}. We use greedy decoding with temperature $0$, do not set top-$p$ explicitly and use provider's default values, set the maximum completion length to 20{,}000 tokens, use a request timeout of 120 seconds, and disable streaming.

\paragraph{System prompt}
\begin{PromptBox}
You are an expert in chemical kinetics and experimental design.
Your mission is to hypothesize plausible mechanisms, and design a single, maximally informative experiment to help distinguish between the plausible candidate mechanisms.
You may propose 5 new experiments.

Execute the following same steps iteratively.
At the beginning of each turn, before executing any step, check how many iterations are done based on the conversation history.

1. **Hypothesize about plausible mechanisms**
    *   You are provided with the fitting results unless this is the first turn. Based on the fitting results, identify all mechanisms that still have non-negligible support.
    *   Even if some mechanisms appear less likely, do NOT prematurely commit to a single leading hypothesis.
    *   Treat hypothesis diversity as important information to preserve.
2. **Hypothesize optimal experiment for discrimination**
    *   Your goal is NOT to maximize the distinction between the single most likely pair of mechanisms.
    *   Instead, your goal is to reduce uncertainty over the entire set of remaining plausible mechanisms.
    *   Prefer experiments that are informative for many mechanisms simultaneously.
    *   Avoid experiments that would only distinguish the top 1–2 candidates while leaving the others indistinguishable.
    *   Consider how the proposed condition helps differentiate multiple mechanisms at once and why it avoids overly myopic or extreme designs.
    *   Reflect on the history of conditions: avoid repeating similar regimes, but also avoid extreme conditions that collapse mechanistic differences (e.g., saturation, depletion).
3.  **Propose Experiment**:
    *   The experimental conditions that you can propose are initial concentration of substrate, catalyst and product.
    *   Conclude with <suggestion> ... </suggestion> and <<DONE>>.

After you propose a new experiment, you would obtain the fitting result of all runs including new experimental data based on your suggestion.

For each turn, you must follow this format:
<iteration>
Specify the iteration number. Start from 1 and increment by 1 iteration.
</iteration>
<reflection>
- Your observations and reasoning about the previous turn's results.
- A clear statement on whether your aim for the turn was achieved.
</reflection>
<hypothesis>
Your detailed reasoning for the proposed experiment. This should include your scientific hypothesis, how the experiment will distinguish between mechanisms, and why the design is effective.
</hypothesis>
<candidate_mechanisms>
The current plausible candidate mechanisms under consideration. Provide them comma-separated as a list (e.g. M1, M2).
</candidate_mechanisms>
<suggestion>
{
    "initial_concentration_of_substrate": <float 0.1 - 5.0>,
    "initial_concentration_of_catalyst": <float 5e-4 - 0.20>,
    "initial_concentration_of_product": <float 0.0 - 0.5>,
}
</suggestion>
<<DONE>>

# 20 Candidate Mechanisms
[MECHANISM-DEFINITION JSON OMITTED FOR BREVITY.]
# Data Structure
The following describes the structure of the data. This is not provided to you directly, but you can access the fitting results of all runs through the conversation history.
At the first run, there is only "run_1". As you propose new experiments and obtain new data, additional runs (e.g., "run_2", "run_3") will be added.

```python
{
  "run_1": {
    "initial_concentration_of_catalyst": float, # Example: 0.01
    "time_data": list[float], # List of time points
    "product_data": list[float], # List of product concentrations at each time point
    "substrate_data": list[float] # List of substrate concentrations at each time point
  },
  "run_2": {
    # ... same structure as run_1
  },
  # ... additional runs
}
```

# Fitting Results
Everytime you propose a new experiment, you will receive the fitting results for all runs including the new experimental data based on your suggestion.
The fitting results are provided as negative log-likelihood values for each candidate mechanism.
Here are example fitting results from the previous runs:
```json
{
    "M1": -66.4739509769402,
    "M2": -66.18465035678079,
    "M3": -118.16423286937294,
    "M4": -131.86593439082938,
    "M5": 35.36495497243471,
    "M6": -78.6224862051261,
    "M7": -77.53127516407199,
    "M8": 69.72349746655837,
    "M9": -93.85551875223946,
    "M10": -106.0115552520685,
    "M11": -111.90660321199617,
    "M12": -117.2823521990672,
    "M13": -127.68391782844185,
    "M14": -66.05561653283398,
    "M15": -135.0936662137469,
    "M16": -71.82102553018892,
    "M17": -66.07488048869106,
    "M18": -63.6419421066368,
    "M19": -88.57669991706247,
    "M20": -48.13563459951533
}
```
\end{PromptBox}
\noindent The omitted mechanism-definition JSON is provided in Appendix~\ref{app:mechanism_definition_json}.

\section{Baselines and tool-based reference}
\label{app:baselines}
All baselines and reference optimize over the same three-dimensional experimental condition space: initial substrate concentration $S \in [0.1, 5.0]$, product concentration $P \in [0, 0.5]$, and catalyst concentration $\mathrm{cat} \in [5 \times 10^{-4}, 0.2]$. Sampling is log-uniform in each coordinate. For the zero lower bound of $P$, we use an epsilon-shifted log-uniform transform with $\epsilon = 10^{-3}$.

\subsection{Maximin Latin hypercube sampling non-adaptive baseline}
The Maximin LHS baseline is non-adaptive. For each run assignment, it generates a 5-point Latin hypercube design over $(S, P, \mathrm{cat})$, maps each coordinate to the log-uniform concentration ranges above, and repeats this for 128 random LHS trials. It selects the design with the largest minimum Euclidean pairwise distance in log-concentration space. The selected 5 points are then used sequentially across the 5 experiment steps. The seed is
\[
42 + 1{,}000{,}003 \times \mathrm{assignment\_id}.
\]

\subsection{Bayesian optimization adaptive baseline}
The Bayesian optimization baseline uses Bayesian optimization to minimize the mechanism entropy computed from NLL feedback. The first two experiments are LHS initialization points. It fits a Gaussian process from log-concentrations to mechanism entropy and samples 512 log-uniform candidate conditions. The acquisition function is the GP lower confidence bound,
\[
a(x) = \mu(x) - \beta \sigma(x),
\]
with $\beta = 1.0$; the condition with minimum $a(x)$ is selected. The GP uses a
\[
\mathrm{ConstantKernel} \times \mathrm{RBF} + \mathrm{WhiteKernel}
\]
kernel, target normalization, $\alpha = 10^{-6}$, and 2 optimizer restarts.

\subsection{Tool-based frozen-parameter EIG reference}
The tool-based EIG reference is an adaptive reference that selects experiments by maximizing an approximate expected information gain over candidate mechanisms. At step 1, it uses the first point from a 5-point maximin-LHS design. From step 2 onward, all candidate mechanisms are refit to the cumulative experimental data $D$, and each mechanism is represented by its MAP parameter estimate.

For each candidate condition $x$, we compute
\[
\mathrm{EIG}(x)
=
H\!\left(p(M \mid D)\right)
-
\mathbb{E}_{M \sim p(M \mid D)}
\left[
H\!\left(p(M' \mid D, y_M(x))\right)
\right].
\]
Here, $M$ denotes the mechanism assumed to generate a hypothetical observation, and $M'$ denotes the mechanism being evaluated after observing that hypothetical data. The distribution $p(M \mid D)$ is the current posterior weight over candidate mechanisms given the cumulative data $D$. We approximate it from the fitted negative log likelihoods as
\[
p(M \mid D) = \mathrm{softmax}(-\mathrm{NLL}_M).
\]
The function $H(\cdot)$ denotes Shannon entropy. Thus, the first term,
$H(p(M \mid D))$, is the current uncertainty over mechanisms before running the next experiment. The second term is the expected posterior uncertainty after hypothetically observing data from condition $x$.

To estimate this expectation, each candidate mechanism $M$ is treated in turn as the source mechanism. We simulate its MAP-fitted trajectory under condition $x$ to obtain a synthetic observation $y_M(x)$. We then compare this synthetic trajectory against the MAP-fitted trajectories of all candidate mechanisms $M'$, using a Gaussian trajectory likelihood with noise scale $\sigma = 0.05$. This yields an updated posterior distribution $p(M' \mid D, y_M(x))$ and its entropy. These posterior entropies are averaged over source mechanisms using the prior weights $p(M \mid D)$.

At each adaptive step, MAP-EIG constructs a candidate pool of 100 conditions using maximin-LHS. The maximin-LHS procedure draws 16 independent 100-point LHS designs and retains the design with the largest minimum pairwise distance in log-concentration space. MAP-EIG is evaluated on the retained 100 candidate conditions, and the condition with the largest EIG value is selected.

\section{Metrics and statistics}
\label{app:metrics_stats}

\paragraph{Mechanism entropy}
Mechanism entropy is computed from per-mechanism negative log-likelihood values $\{\mathrm{NLL}_i\}_{i=1}^{|\mathcal{M}|}$ (lower is better) by converting them to a normalized distribution via a softmax over $-\mathrm{NLL}$,
\[
p_i \;=\; \frac{\exp(-\mathrm{NLL}_i)}{\sum_{j=1}^{|\mathcal{M}|}\exp(-\mathrm{NLL}_j)} \, ,
\]
and then computing the (natural-log) Shannon entropy
\[
H \;=\; -\sum_{i=1}^{|\mathcal{M}|} p_i \log p_i \, .
\]
We report 95\% bootstrap confidence intervals over the 200 systems.

\paragraph{Occupancy entropy.}
We define occupancy entropy by discretizing the proposal space into fixed bins and measuring how broadly the cumulative proposals occupy those bins. Each proposal \(\alpha\) consists of substrate concentration \(S\), catalyst concentration \(C\), and product concentration \(P\).

The \(S\) and \(C\) coordinates are first mapped to log space and normalized to \([0,1]\):
\[
u_d =
\frac{\log_{10}(x_d) - \log_{10}(l_d)}
{\log_{10}(h_d) - \log_{10}(l_d)},
\quad d \in \{S,C\},
\]
where \(l_d\) and \(h_d\) are the lower and upper bounds of that dimension. Each of \(S\) and \(C\) is then assigned to one of three equal-width bins in this normalized log space. The product coordinate \(P\) is assigned to one of two bins using the raw-space threshold \(P = 0.1\): one bin for \(P < 0.1\), and one bin for \(P \geq 0.1\). This gives \(3 \times 3 \times 2 = 18\) total proposal bins.

Let \(i\) index the ground-truth sample and let \(\alpha_{i,k}\) denote the proposal made for sample \(i\) at iteration \(k\). For each sample \(i\) and iteration \(t\), we compute the empirical distribution over the 18 bins using the cumulative proposals from iterations 1 through \(t\). If \(p_{i,t,b}\) is the fraction of these proposals assigned to bin \(b\), the sample-level occupancy entropy is
\[
H_{i,t} =
-\frac{1}{\log 18}
\sum_{b=1}^{18} p_{i,t,b} \log p_{i,t,b},
\]
where bins with \(p_{i,t,b}=0\) are omitted from the sum. The normalization by \(\log 18\) makes the maximum possible value equal to 1. The plotted occupancy entropy at iteration \(t\) is the mean of \(H_{i,t}\) over the 200 ground-truth samples.

\paragraph{S-cat corner mass}
We define S-cat corner mass as the fraction of proposed conditions that lie near a corner of the two-dimensional substrate--catalyst design space. For each proposal, we ignore the product concentration and use only the substrate concentration \(S\) and catalyst concentration \(C\). Each coordinate is first mapped to log space and normalized to \([0,1]\):
\[
u_d =
\frac{\log_{10}(x_d) - \log_{10}(l_d)}
{\log_{10}(h_d) - \log_{10}(l_d)},
\quad d \in \{S,C\},
\]
where \(l_d\) and \(h_d\) are the lower and upper bounds of that design dimension. Values are clipped to \([0,1]\).

A proposal \(\alpha\) is counted as an S-cat corner proposal if both normalized coordinates are within \(\tau = 0.15\) of either boundary:
\[
\min(u_S(\alpha), 1-u_S(\alpha)) \leq \tau
\quad \mathrm{and} \quad
\min(u_C(\alpha), 1-u_C(\alpha)) \leq \tau .
\]
Thus all four corners of the \(S\)-\(C\) plane are included.

Let \(i\) index the ground-truth sample and let \(\alpha_{i,k}\) denote the proposal made for sample \(i\) at iteration \(k\). For each sample \(i\) and iteration \(t\), we define \(m_{i,t}\) as the cumulative corner fraction up to iteration \(t\):
\[
m_{i,t} = \frac{1}{t}\sum_{k=1}^{t} I(\alpha_{i,k}),
\]
where \(I(\alpha_{i,k}) = 1\) if proposal \(\alpha_{i,k}\) is an S-cat corner proposal and \(0\) otherwise. The plotted S-cat corner mass at iteration \(t\) is the mean of \(m_{i,t}\) over the 200 ground-truth samples.

\section{Failures and edge cases}
\label{app:failures_and_edge_cases}
If the fitting procedure fails to return valid NLL values for all 20 mechanisms for a proposed condition $\alpha$ (e.g., due to numerical instability or divergence in ODE simulation), then in NLL mode we return \texttt{null} NLL values to the agent for that iteration. For all subsequent iterations, we exclude the corresponding observation $O(\alpha)$ from the accumulated collection of kinetic time-series observations $\mathcal{O}_{1:k}$ used for fitting and for computing evaluation metrics (accuracy and entropy).
In the agent loop, we additionally provide an environment message indicating that the previous condition diverged and requesting a different proposal. For the random baseline, no such message is required; we simply proceed with the next sampled condition under the same exclusion rule.

\section{Hypothesis-driven reasoning analysis}
\subsection{The relation between the prompt instruction and the evaluation criteria}
In the benchmark, the agent prompt (Appendix \ref{app:agent_prompt}) provides only high-level objectives. We analyze whether agents exhibit these intermediate reasoning steps when high-level instructions are provided. Table \ref{tab:capabilities} summarizes the relation between prompt instructions and evaluation criteria.

\begin{table}[h]
\centering
\caption{The four reasoning steps evaluated in our benchmark. The agent prompt (Appendix~\ref{app:agent_prompt}) provides only high-level objectives without prescribing the intermediate reasoning steps required to achieve them. The rightmost column describes the reasoning operation that connects the prompt-level objective to the performance required at the highest score level.}
\label{tab:capabilities}
\footnotesize
\setlength{\tabcolsep}{4pt}
\begin{tabular}{p{1.7cm} p{4.8cm} p{3.0cm} p{3.0cm}}
\toprule
\textbf{Step} & \textbf{What is evaluated} & \textbf{Prompt instruction} & \textbf{Intermediate reasoning operation} \\
\midrule
Belief update &
Whether the candidate mechanisms stated in the trace are consistent with the per-mechanism NLL values returned in the previous turn. The highest score is given when the candidate set or ranking accurately reflects the top candidate cluster implied by the NLL feedback. &
\emph{``identify all mechanisms that still have non-negligible support''} (goal-level) &
Translating quantitative likelihood signals into a correctly ordered candidate set \\
\addlinespace
Hypothesis discrimination &
Whether the trace articulates mechanistically meaningful differences among the top candidate mechanisms. The highest score is given when discriminative mechanistic features are identified at the level of sub-reaction steps, catalyst states, reversibility, or active species, rather than at coarse family levels. &
No explicit instruction; implicit prerequisite of predictive reasoning and discriminative experiment design &
Articulating mechanistic differences at sub-reaction-step resolution rather than family level \\
\addlinespace
Discriminative experiment design &
Whether the agent leverages its own identified mechanistic features when selecting a concrete experimental condition. The highest score is given when the proposal directly and intentionally probes the specific mechanistic feature identified in the reasoning. &
\emph{``Prefer experiments that are informative for many mechanisms simultaneously''} (goal-level) &
Operationalizing the identified mechanistic features as an actionable experimental condition \\
\addlinespace
Predictive reasoning &
Whether the trace predicts contrastively how the proposed experiment would behave under different candidate mechanisms. The highest score is given when predictions are mechanism-specific and tied to observable consequences such as curvature, early slowdown, or final conversion. &
\emph{``Consider how the proposed condition helps differentiate multiple mechanisms''} (goal-level) &
Translating identified mechanistic features into sub-reaction-step-level outcome predictions \\
\bottomrule
\end{tabular}
\end{table}

\subsection{Hypothesis-driven reasoning steps}
\label{app:hypothesis_driven_reasoning_capabilities}
\textbf{i) Belief update step}

We evaluate whether the trace's stated candidate mechanisms are aligned with the immediately preceding NLL feedback. This measures whether the model is tracking the current hypothesis space correctly, not whether it eventually identifies the correct mechanism.

\textbf{Score 0 (Misaligned).}
The trace fails to mention the relevant top mechanisms, or discusses mechanisms that are clearly inconsistent with the NLL feedback.

\textbf{Score 1 (Partially aligned).}
The trace mentions some relevant candidates, but important top mechanisms are missing, misranked, or replaced by implausible alternatives.

\textbf{Score 2 (Aligned).}
The trace accurately reflects the top candidate set or top candidate cluster implied by the NLL feedback.

\textbf{ii) hypothesis discrimination step}

We evaluate whether the trace identifies the mechanistically relevant differences among the current top candidate mechanisms.

\textbf{Score 0}
The trace does not identify meaningful mechanistic differences among the top candidates.

\textbf{Score 1} 
The trace identifies only coarse family-level distinctions, such as activation vs. deactivation or dimerization vs. product effect.

\textbf{Score 2}
The trace identifies discriminative mechanistic features at the level of sub-reaction steps, catalyst states, reversibility, active species, or other specific mechanistic motifs.
            
\textbf{iii) predictive reasoning step}
We evaluate whether the trace predicts how the proposed experiment would behave under different candidate mechanisms.

\textbf{Score 0}
The trace proposes an experiment without giving a concrete prediction of how different candidate mechanisms would respond.

\textbf{Score 1}
The trace gives a prediction, but only at a coarse family level or in vague terms such as 'should increase rate' or 'should show inhibition'.

\textbf{Score 2}
The trace gives contrastive, mechanism-specific predictions tied to sub-reaction steps, species, equilibria, rate limitation, or clearly stated observable consequences such as curvature, early slowdown, or final conversion.

\textbf{iv) discriminative experiment design step}

We evaluate whether the proposed experiment directly tests the mechanistic difference identified in the reflection or hypothesis.

\textbf{Score 0}
The proposal does not meaningfully probe the mechanistic feature that the trace claims is important.

\textbf{Score 1}
The proposal is loosely related to the claimed feature, but remains generic, indirect, or only partially diagnostic.

\textbf{Score 2}
The proposal directly and intentionally probes the specific mechanistic feature identified in the reasoning.

\subsection{LLM-as-a-Judge Prompt}
\label{app:judge_prompt}

We evaluated each reasoning trace independently along each rubric axis. For each trace-rubric pair, the judge received a fixed system prompt and a rubric-specific user prompt. The rubric definitions and score descriptions are written in \ref{app:hypothesis_driven_reasoning_capabilities}
. The trace payload contained the model's trace, mechanism definitions, and the likelihood feedback at the previous iteration. The judge was instructed to score only the requested axis and to return strict JSON.
\paragraph{System prompt.}
\begin{PromptBox}
You are evaluating one rubric dimension of scientific reasoning quality.

You must judge only the requested rubric axis.
Use only the information provided in the trace payload and rubric instructions.
Do not use outside knowledge of the ground truth beyond what is explicitly given.
Return strict JSON only.
\end{PromptBox}

\paragraph{User prompt template.}
For each rubric axis, we instantiated the following template with the corresponding rubric definition, scoring scale, focus fields, and trace payload.

\begin{PromptBox}
Task:
Score the trace on the rubric dimension "{rubric_display_name}".

Definition:
{rubric_definition}

Scoring rubric:
- 0: {score_0_description}
- 1: {score_1_description}
- 2: {score_2_description}

Focus your judgment on these parts of the trace:
- {field_1}
- {field_2}
- ...

Do not reward:
- {do_not_reward_1}
- {do_not_reward_2}
- ...

Additional notes:
- {note_1}
- {note_2}
- ...

Return JSON with exactly these fields:
{
  "score": 0 or 1 or 2,
  "confidence": 0.0 to 1.0,
  "evidence": ["short quote or tight paraphrase 1", "short quote or tight paraphrase 2"],
  "rationale": "1-3 sentences explaining why the evidence maps to the rubric score"
}

Trace payload:
{trace_payload_json}
\end{PromptBox}

\subsection{Reliability of LLM-as-a-Judge Evaluation}
\label{app:judge_agreement}

We assess the reliability of LLM-judge scores in two stages using standard-prompt traces only; these human annotations do not directly validate scores on four-step-prompt traces. First, one author annotates 30 reasoning traces across all four rubric dimensions, providing an initial assessment of author-LLM agreement. We then extend the evaluation to 60 reasoning traces annotated by two researchers with chemistry expertise, focusing on discriminative experiment design and predictive reasoning. The annotators were labmates who volunteered their assistance without compensation and scored the traces independently. They used the rubric in Appendix~\ref{app:hypothesis_driven_reasoning_capabilities}; the annotation interface and instructions are included with the released code. The authors considered formal ethics review unnecessary for this voluntary assessment of model outputs, and no formal review was obtained. These dimensions are selected because they exhibit substantial cross-model variation and strong correlations with task accuracy in the initial LLM-judge analysis. For both evaluations, we report agreement at the trace and model-aggregated levels.

Our main diagnostic claim concerns differences among the evaluated models: predictive reasoning and discriminative experiment design vary across models and correlate with task accuracy. Interpreting these associations as planning bottlenecks requires validating the scores, with stronger independent support for PR than for DED in this setting. We therefore assess agreement at the model level, while also examining trace-level agreement to evaluate the reliability of individual scores. We interpret individual trace scores as discrete and noisy operationalizations of underlying reasoning quality rather than exact ground-truth labels. Under a coarse 0/1/2 rubric, especially for borderline traces, some loss of agreement at the trace level is therefore expected, while model-aggregated scores recover a more stable signal.

For discriminative experiment design and predictive reasoning, we assess three aspects of agreement: (i) whether model-level human–LLM Spearman correlations reach our strong-agreement criterion of \(\rho \geq 0.70\); (ii) how these correlations compare with human–human agreement at the model level; and (iii) how human–LLM agreement compares with human–human agreement at the trace level. Because the initial author annotations are provided by a single annotator, they support assessment of criterion (i), but not the comparisons with human-human agreement in (ii) and (iii). We report author-LLM correlations at both the model and trace level. For the annotations from the two independent raters, we assess all three aspects. Thirty of the 60 expert-rated traces also occur in the author-calibration set; the expert ratings were collected separately, and author ratings are not pooled with them as independent validation.

\paragraph{Author-LLM agreement}

Table \ref{tab:author_llm_agreement} summarizes author-LLM agreement at the model-aggregated and trace level, respectively. Model-level author-LLM Spearman's $\rho$ for discriminative experiment design and predictive reasoning are in the strong range, exceeding 0.70 for both axes and thus satisfying criterion (i).
Per-trace Spearman's $\rho$ is 0.429 and 0.675 for discriminative experiment design and predictive reasoning, respectively, lower than model-level agreement, as expected. For the saturated axes (belief update, hypothesis discrimination; see Table \ref{tab:author_llm_trace_error}), rank-based statistics at either level are uninformative due to limited variance; we therefore emphasize item-level absolute agreement (Table \ref{tab:author_llm_trace_error}, e.g., mean absolute error $\leq$ 0.25 on a 0–2 scale for both saturated axes).

\begin{table}[h]
\centering
\caption{Author--LLM agreement at the trace and model levels.
LLM-judge scores are the mean of GPT-5.4 and Claude Opus 4.7
scores on the same traces. Model-level scores are averaged
over three traces per model. Brackets indicate 95\% confidence
intervals for Spearman's $\rho$. Trace-level and model-level analyses use 30 traces and 10 models, respectively.}
\label{tab:author_llm_agreement}
\small
\begin{tabular}{llc}
\toprule
Capability & Level & Author--LLM \\
\midrule
Belief Update & Trace & 0.282 \\
& & [-0.161, 0.654] \\
& Model & 0.323 \\
& & [-0.527, 0.893] \\
\addlinespace
Hypothesis Discrimination & Trace & 0.628 \\
& & [0.262, 0.908] \\
& Model & 0.845 \\
& & [0.548, 1.000] \\
\addlinespace
Discriminative Experiment Design & Trace & 0.429 \\
& & [0.080, 0.745] \\
& Model & 0.746 \\
& & [0.271, 0.955] \\
\addlinespace
Predictive Reasoning & Trace & 0.675 \\
& & [0.421, 0.871] \\
& Model & 0.850 \\
& & [0.472, 0.958] \\
\bottomrule
\end{tabular}
\end{table}

\begin{table}[h]
\centering
\caption{Trace-level absolute agreement between author judgments and LLM-as-judge scores. Bias is LLM-as-judge minus author score.}
\label{tab:author_llm_trace_error}
\setlength{\tabcolsep}{5pt}
\begin{tabular}{lrrrrrr}
\toprule
Capability & $n$ & MAE & RMSE & Bias & Within 0.25 & Within 0.5 \\
\midrule
Overall & 120 & 0.258 & 0.456 & 0.075 & 0.642 & 0.842 \\
Belief Update & 30 & 0.250 & 0.456 & 0.150 & 0.667 & 0.833 \\
Hypothesis Discrimination & 30 & 0.167 & 0.365 & 0.100 & 0.767 & 0.900 \\
Discriminative Experiment Design & 30 & 0.317 & 0.508 & 0.050 & 0.567 & 0.800 \\
Predictive Reasoning & 30 & 0.300 & 0.483 & 0.000 & 0.567 & 0.833 \\
\bottomrule
\end{tabular}
\end{table}

\paragraph{Expert-LLM agreement}

Table \ref{tab:expert_llm_agreement} shows expert-expert, expert-LLM agreement at both the model and trace levels. Predictive reasoning meets our prespecified model-level criterion for strong expert–LLM agreement ($\rho \geq 0.70$) for both independent annotators ($\rho =$ 0.862 and 0.920). These estimates are close to the model-level expert–expert agreement ($\rho =$ 0.940). At the trace level, expert–LLM agreement ($\rho =$ 0.590 and 0.723) is also similar to expert–expert agreement ($\rho =$ 0.651). Thus, predictive reasoning satisfies all three of our stated validation criteria.

For discriminative experiment design, one independent annotator shows strong agreement with the LLM judge ($\rho =$ 0.820, above our prespecified threshold of 0.70), whereas the other shows moderate agreement ($\rho =$0.579) at the model level. The expert–LLM point estimates are comparable to or higher than the expert–expert estimate ($\rho =$0.596). However, the expert–expert model-level estimate has a wide 95\% CI that includes zero (95\% CI [-0.264, 0.913]), so this comparison does not provide a precise model-level validation of expert–LLM agreement. At the trace level, expert–LLM agreement is positive ($\rho =$0.407 and 0.535), but lower in point estimate than expert–expert agreement ($\rho =$ 0.727), although the bootstrap confidence intervals overlap. Overlap alone does not establish equivalent agreement. Thus, the independent evidence for discriminative experiment design is positive but less consistent than for PR, particularly at the trace level.

Overall, predictive reasoning meets our stated validation criteria within the evaluated setting. For discriminative experiment design, they show positive expert–LLM agreement and encouraging model-level evidence, but do not validate the diagnostic as consistently as predictive reasoning, particularly at the trace level. We therefore interpret the association between LLM-judged discriminative experiment design scores and task accuracy as suggestive evidence of a reasoning bottleneck.

\begin{table}[h]
\centering
\caption{Expert--expert and expert--LLM agreement for discriminative
experiment design and predictive reasoning.
Values are Spearman's $\rho$ with 95\% bootstrap confidence intervals.
Trace-level and model-level analyses use 60 traces and 10 models,
respectively.}
\label{tab:expert_llm_agreement}
\small
\setlength{\tabcolsep}{4pt}
\begin{tabular}{llccc}
\toprule
Axis & Level & Expert--expert & Annotator 1--LLM & Annotator 2--LLM \\
\midrule
Discriminative experiment design & Trace
& 0.727 & 0.407 & 0.535 \\
& & [0.510, 0.901] & [0.142, 0.631] & [0.304, 0.710] \\
\addlinespace
Discriminative experiment design & Model
& 0.596 & 0.579 & 0.820 \\
& & [$-0.264$, 0.913] & [$-0.121$, 0.875] & [0.537, 0.974] \\
\midrule
Predictive reasoning & Trace
& 0.651 & 0.590 & 0.723 \\
& & [0.445, 0.810] & [0.360, 0.763] & [0.552, 0.850] \\
\addlinespace
Predictive reasoning & Model
& 0.940 & 0.862 & 0.920 \\
& & [0.757, 1.000] & [0.459, 0.991] & [0.634, 0.997] \\
\bottomrule
\end{tabular}
\end{table}

\paragraph{Expert assessment of extreme LLM scores}

As a supplementary analysis of the same 60 expert-annotated traces reported above, we examined expert ratings for cases receiving extreme LLM-judge scores. 
This analysis reuses the existing expert annotations and does not constitute an additional independent validation set. To require consensus from two non-self judges, we exclude the 12 traces generated by GPT-5.4 or Claude Opus 4.7 before selecting extreme scores, leaving 48 eligible traces from eight models.
Table \ref{tab:expert_extreme_llm_scores} summarizes expert assessment of extreme LLM-judge scores.
For predictive reasoning, the three traces unanimously scored 0 by the two LLM judges received mean expert scores of 0.33 and 0.00, whereas the 17 traces unanimously scored 2 received means of 1.82 and 1.94. 
For discriminative experiment design, the 21 traces unanimously scored 2 received a mean of 1.95 from each expert. No trace in this sample received a unanimous LLM score of 0 for discriminative experiment design. These results support expert alignment at the high end of both dimensions and provide limited evidence of alignment at the low end of predictive reasoning. The low end of discriminative experiment design remains insufficiently assessed.

\begin{table}[h]
\centering
\caption{
Expert assessment of extreme LLM-judge scores within the same
60 expert-annotated traces used in the agreement analysis.
Experts were blinded to generating-model identities, LLM scores,
and judge rationales. Consensus selection excludes traces generated by either judge model, leaving 48 eligible traces from the 60-trace annotation set.
Parentheses indicate exact-match rates.
No trace received a unanimous discriminative experiment design score of 0.
}
\label{tab:expert_extreme_llm_scores}
\small
\setlength{\tabcolsep}{5pt}
\begin{tabular}{llrcc}
\hline
Axis & LLM-judge scores & $n$ &
\shortstack{Expert 1 mean\\(exact match)} &
\shortstack{Expert 2 mean\\(exact match)} \\
\hline
Predictive reasoning  & Both judges: 0       & 3  & 0.333 (66.7\%) & 0.000 (100.0\%) \\
Predictive reasoning  & Both judges: 2       & 17 & 1.824 (88.2\%) & 1.941 (94.1\%) \\
\hline
Discriminative experiment design & Both judges: 0       & 0  & ---            & ---             \\
Discriminative experiment design & Both judges: 2       & 21 & 1.952 (95.2\%) & 1.952 (95.2\%) \\
\hline
\end{tabular}
\end{table}

\paragraph{Judge--judge agreement and self-preference}
On the 400 iteration-3 traces (40 per model), the two judges have model-level Spearman correlations of 0.982 for DED (95\% bootstrap CI [0.867, 1.000]) and 0.963 for PR ([0.796, 1.000]). Trace-level correlations are 0.545 ([0.467, 0.619]) and 0.768 ([0.715, 0.820]), respectively. This consistency check complements rather than replaces independent expert validation. These judge--judge statistics are distinct from the pooled reasoning--accuracy correlations of 0.918 and 0.736 reported in the main analysis.

Table~\ref{tab:self_preference_exclusion} reports the iteration-3 reasoning--accuracy correlations under three aggregation rules: averaging both judges, excluding each judge's scores of its own model, and excluding each judge's scores of models from the same provider. The associations are similar under these exclusions. The direct mean difference, GPT-judge minus Claude-judge score across all four axes, is +0.038 on GPT-5.4 traces and -0.006 on Claude Opus 4.7 traces. Both signs favor the respective judge's own model; their magnitudes are small on the 0--2 scale, but without uncertainty intervals they do not establish either the presence or absence of systematic self-preference. We did not evaluate a third disjoint judge.

\begin{table}[ht]
\centering
\caption{Iteration-3 model-level Spearman correlations between reasoning scores and final accuracy under judge-exclusion rules. Brackets indicate 95\% bootstrap confidence intervals over ten models. The main pooled analysis uses same-model exclusion.}
\label{tab:self_preference_exclusion}
\small
\begin{tabular}{lcc}
\toprule
Judge aggregation & DED & PR \\
\midrule
Both judges & 0.924 [0.660, 1.000] & 0.778 [0.293, 0.987] \\
Same-model exclusion & 0.924 [0.660, 1.000] & 0.778 [0.293, 0.987] \\
Same-provider exclusion & 0.924 [0.660, 1.000] & 0.766 [0.291, 0.962] \\
\bottomrule
\end{tabular}
\end{table}

\subsection{Robustness across iterations}
Our main analysis aggregates reasoning scores across iterations 2-4. To assess consistency across iterations, we also compute model-level correlations between reasoning scores and final accuracy separately for each iteration (Table \ref{tab:iteration_robustness}). Both discriminative experiment design and predictive reasoning show positive correlations with task accuracy at all three iterations, with all 95\% bootstrap confidence intervals excluding zero. This consistent pattern indicates that the associations observed in the pooled analysis are not confined to a single iteration. Iteration 3 here uses the judging pass collected for the pooled iterations 2--4 analysis. The historical iteration-3 exclusion analysis above uses the earlier pass, so its coefficients differ.

\begin{table}[h]
\centering
\caption{Model-level Spearman correlations between reasoning scores
and final task accuracy, computed separately for iterations 2--4
with self-model exclusion. Brackets indicate 95\% bootstrap
confidence intervals.}
\label{tab:iteration_robustness}
\small
\begin{tabular}{lccc}
\toprule
Reasoning dimension & Iteration 2 & Iteration 3 & Iteration 4 \\
\midrule
Discriminative experiment design
& 0.912 & 0.942 & 0.851 \\
& [0.601, 1.000] & [0.686, 1.000] & [0.488, 0.987] \\
\addlinespace
Predictive reasoning
& 0.693 & 0.809 & 0.827 \\
& [0.092, 0.887] & [0.334, 1.000] & [0.440, 0.991] \\
\bottomrule
\end{tabular}
\end{table}

\section{Prompt ablations}
\subsection{Explicit four-step reasoning instructions}
\label{app:reasoning_step_instruction_ablation}
In the section \ref{sec:experiments}, we observed that weaker models can update beliefs and discriminate among hypotheses, yet struggle to translate this understanding into discriminative experimental designs and sharp predictions. To distinguish limitations of the base model from potential underspecification of the original prompt, we conducted an ablation with a prompt that explicitly instructs the four reasoning steps.

We compare LLM-judged reasoning traces from iterations 2--4 for nine models, matching standard and four-step traces by model, reaction system, and iteration. Two DeepSeek V3.2 iteration-4 four-step traces (M7\_1 and M13\_1) could not be retrieved from Weave. We exclude their standard-prompt counterparts from the paired comparison, leaving 1,078 matched pairs: 118 for DeepSeek and 120 for each of the other eight models. The remaining iterations for these two systems are retained. Mistral Large 3 was omitted because the synchronous OpenRouter endpoint configured for the original experiments became unavailable before the submission period.

\paragraph{Prompt with four-step reasoning step instruction}
\begin{PromptBox}
You are an expert in chemical kinetics and experimental design.
Your mission is to hypothesize plausible mechanisms, and design a single, maximally informative experiment to help distinguish between the plausible candidate mechanisms.
You may propose 5 new experiments.

Execute the following four reasoning steps iteratively before proposing an experiment.
At the beginning of each turn, before executing any step, check how many iterations are done based on the conversation history.

1. **Belief update**
    * You are provided with fitting results unless this is the first turn.
    * State the candidate mechanisms or candidate cluster that remain plausible according to the latest NLL feedback. Preserve relevant rankings and near-ties.

2. **Hypothesis discrimination**
    * Identify the specific mechanistic differences among the leading candidates that are most useful to test next.
    * Name the relevant sub-reaction step, catalyst state, active or inactive species, reversibility, equilibrium, saturation regime, or rate-limiting process.

3. **Discriminative experiment design**
    * Propose one experiment that directly tests the mechanistic feature identified in Step 2.
    * Prefer an experiment that separates multiple remaining candidates, rather than one that only distinguishes the top one or two candidates.
    * Avoid repeating similar conditions or choosing extreme conditions that collapse mechanistic differences, such as complete saturation or depletion.

4. **Contrastive prediction**
    * For the exact condition proposed in Step 3, state contrasting predictions for each leading candidate or candidate cluster.
    * Explain how each prediction follows from the mechanistic feature in Step 2.
    * State observable consequences explicitly—for example, initial-rate dependence, curvature, early slowdown, saturation, inhibition profile, or final conversion.
    * Do not use vague statements such as “the rate should change” without specifying how the outcomes differ across candidates.

5.  **Propose Experiment**:
    *   The experimental conditions that you can propose are initial concentration of substrate, catalyst and product.
    *   Conclude with <suggestion> ... </suggestion> and <<DONE>>.

After you propose a new experiment, you would obtain the fitting result of all runs including new experimental data based on your suggestion.

For each turn, you must follow this format:
<iteration>
Specify the iteration number. Start from 1 and increment by 1 iteration.
</iteration>
<reflection>
- Your observations and reasoning about the previous turn's results.
- A clear statement on whether your aim for the turn was achieved.
</reflection>
<hypothesis>
Your detailed reasoning for the proposed experiment. This should include your thoughts about 2. hypothesis discrimination, 3. discriminative experiment design and 4. contrastive prediction.
</hypothesis>
<candidate_mechanisms>
List the plausible candidate mechanisms, sorted in descending order of plausibility (most likely first). Provide them as a JSON list of strings (e.g. ["M1", "M2"]).
</candidate_mechanisms>
<suggestion>
{{
    "initial_concentration_of_substrate": <float 0.1 - 5.0>,
    "initial_concentration_of_catalyst": <float 5e-4 - 0.20>,
    "initial_concentration_of_product": <float 0.0 - 0.5>,
}}
</suggestion>
<<DONE>>

# 20 Candidate Mechanisms
[MECHANISM-DEFINITION JSON OMITTED FOR BREVITY.]

# Data Structure
The following describes the structure of the data. This is not provided to you directly, but you can access the fitting results of all runs through the conversation history.
At the first run, there is only "run_1". As you propose new experiments and obtain new data, additional runs (e.g., "run_2", "run_3") will be added.

```python
{{
  "run_1": {{
    "initial_concentration_of_catalyst": float, # Example: 0.01
    "time_data": list[float], # List of time points
    "product_data": list[float], # List of product concentrations at each time point
    "substrate_data": list[float] # List of substrate concentrations at each time point
  }},
  "run_2": {{
    # ... same structure as run_1
  }},
  # ... additional runs
}}
```

# Fitting Results
Everytime you propose a new experiment, you will receive the fitting results for all runs including the new experimental data based on your suggestion.
The fitting results are provided as negative log-likelihood values for each candidate mechanism.
Here are example fitting results from the previous runs:
```json
{{
    "M1": -66.4739509769402,
    "M2": -66.18465035678079,
    "M3": -118.16423286937294,
    "M4": -131.86593439082938,
    "M5": 35.36495497243471,
    "M6": -78.6224862051261,
    "M7": -77.53127516407199,
    "M8": 69.72349746655837,
    "M9": -93.85551875223946,
    "M10": -106.0115552520685,
    "M11": -111.90660321199617,
    "M12": -117.2823521990672,
    "M13": -127.68391782844185,
    "M14": -66.05561653283398,
    "M15": -135.0936662137469,
    "M16": -71.82102553018892,
    "M17": -66.07488048869106,
    "M18": -63.6419421066368,
    "M19": -88.57669991706247,
    "M20": -48.13563459951533
}}
```
\end{PromptBox}

\begin{figure}[h]
    \centering
    \includegraphics[width=0.9\linewidth, trim=0cm 0.0cm 0cm 0.3cm]{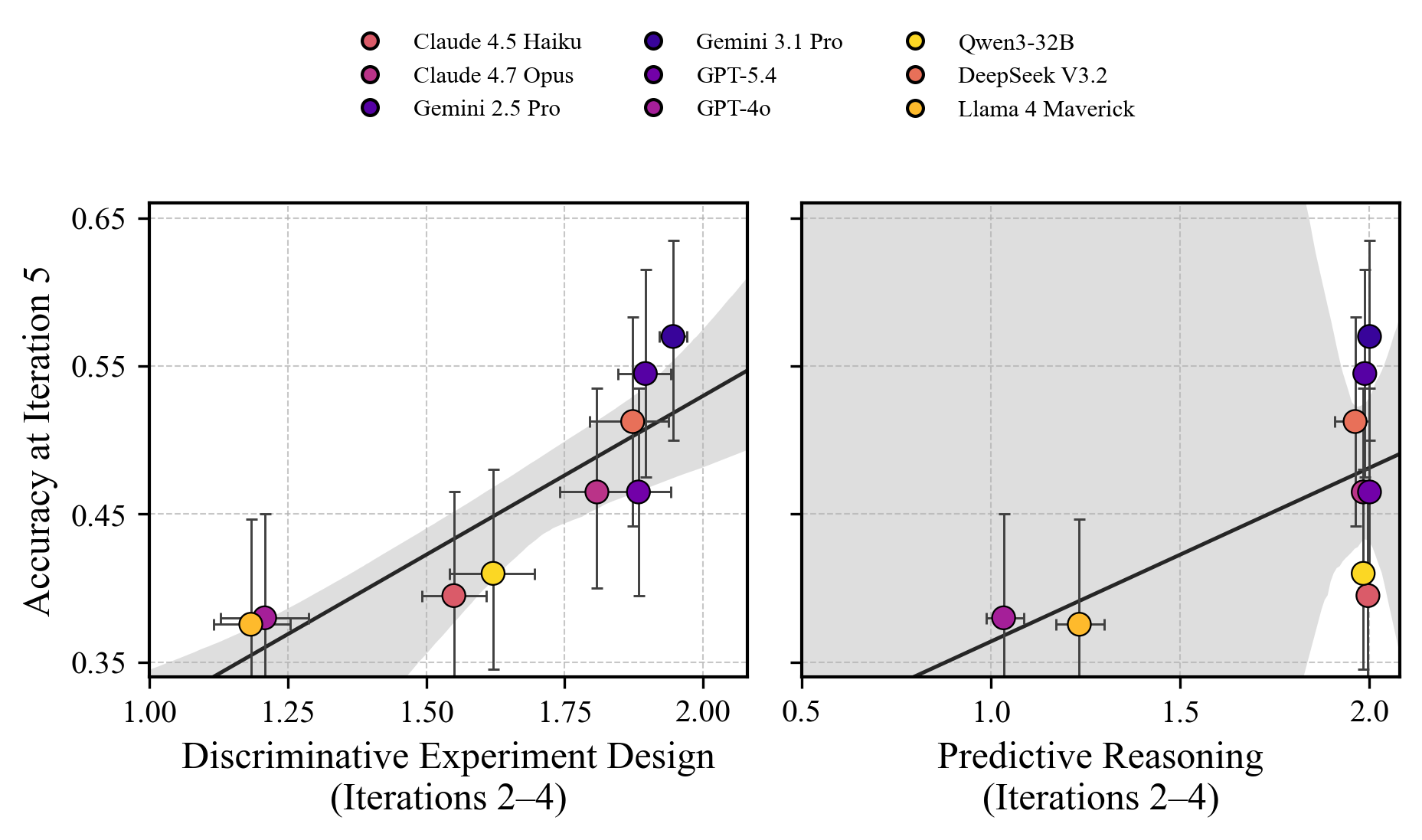}
    \caption{
    Model-aggregated reasoning scores pooled across iterations 2--4 are plotted against task accuracy at iteration 5 for the four-step instruction ablation. Each point denotes a model; horizontal and vertical error bars indicate 95\% bootstrap confidence intervals. Solid lines show ordinary least-squares fits, and shaded regions show 95\% bootstrap confidence bands. Discriminative experiment design exhibits a strong positive association with task accuracy ($\rho=0.971$, 95\% bootstrap CI $[0.762, 1.000]$), whereas predictive reasoning (PR) has a weaker, uncertain association ($\rho=0.591$, 95\% bootstrap CI $[-0.193, 0.930]$).
    }
    \label{fig:reasoning_task_accuracy_correlation_four_step_instruction_ablation}
\end{figure}

Figure \ref{fig:paired_effects_four_step_prompting} compares the standard and four-step prompts across nine models using matched traces pooled across iterations 2--4. The four-step instruction increases predictive-reasoning point estimates for all nine models, whereas changes in discriminative experiment design are smaller and more heterogeneous.
Figure \ref{fig:reasoning_task_accuracy_correlation_four_step_instruction_ablation} plots each model's discriminative experiment design and predictive reasoning scores, pooled across iterations 2--4, against task accuracy under the four-step instruction prompt.
Across models under the four-step condition, discriminative experiment design is strongly associated with task accuracy (Spearman's \(\rho=0.971\), bootstrap 95\% CI \([0.762, 1.000]\)), while the association for predictive reasoning is positive but uncertain (\(\rho=0.591\), 95\% CI \([-0.193, 0.930]\)).

Table~\ref{tab:step_prompt_accuracy} compares identification accuracy under the two prompts across all nine models, including paired changes and their uncertainty. Although predictive-reasoning scores increased under the four-step prompt for all nine models, every paired 95\% accuracy-change interval includes zero; the results therefore establish neither an accuracy improvement nor equivalence between prompts. This suggests that eliciting more explicit predictions alone may be insufficient to improve experiment selection. The observed association between discriminative experiment design and accuracy motivates further investigation into interventions that improve the design of informative experiments.

\begin{table}[ht]
\centering
\caption{Task accuracy under standard and four-step prompts. The sample-count column gives standard/four-step counts of systems with a valid cumulative fit by iteration 5. Paired changes and 95\% bootstrap intervals use shared valid systems; their point estimates can differ from the difference of the full-condition means.}
\label{tab:step_prompt_accuracy}
\small
\setlength{\tabcolsep}{4pt}
\begin{tabular}{lcccc}
\toprule
Model & Standard & Four-step & $n$ & Paired change [95\% CI] \\
\midrule
Gemini 3.1 Pro & 0.610 & 0.570 & 200/200 & -0.040 [-0.120, 0.035] \\
Gemini 2.5 Pro & 0.575 & 0.545 & 200/200 & -0.030 [-0.115, 0.055] \\
Claude Opus 4.7 & 0.520 & 0.465 & 200/200 & -0.055 [-0.135, 0.030] \\
GPT-5.4 & 0.470 & 0.465 & 200/200 & -0.005 [-0.095, 0.080] \\
DeepSeek V3.2 & 0.470 & 0.513 & 200/199 & +0.045 [-0.040, 0.136] \\
GPT-4o & 0.420 & 0.380 & 200/200 & -0.040 [-0.115, 0.035] \\
Claude 4.5 Haiku & 0.435 & 0.395 & 200/200 & -0.040 [-0.120, 0.040] \\
Llama 4 Maverick & 0.425 & 0.376 & 200/197 & -0.041 [-0.132, 0.046] \\
Qwen3-32B & 0.395 & 0.410 & 200/200 & +0.015 [-0.065, 0.095] \\
\bottomrule
\end{tabular}
\end{table}

Across all nine models, changes in exploration are heterogeneous (Table~\ref{tab:step_prompt_exploration}). Gemini 2.5 Pro shows higher occupancy entropy, whereas Claude Opus 4.7, GPT-4o, and DeepSeek V3.2 show lower occupancy entropy. Claude 4.5 Haiku and DeepSeek V3.2 increase S--cat corner mass, whereas Gemini 2.5 Pro decreases it. The remaining intervals include zero. Comparisons use shared systems with a recorded fifth proposal; incomplete proposal histories produce different sample counts and limit conclusions about missing runs.

\begin{table}[ht]
\centering
\caption{Paired changes in exploration at iteration 5, four-step minus standard. Brackets give exploratory 95\% bootstrap intervals over shared systems with recorded fifth proposals.}
\label{tab:step_prompt_exploration}
\small
\setlength{\tabcolsep}{4pt}
\begin{tabular}{lccc}
\toprule
Model & $n$ & $\Delta$ occupancy entropy & $\Delta$ S--cat corner mass \\
\midrule
Gemini 3.1 Pro & 198 & -0.004 [-0.015, 0.008] & +0.002 [-0.029, 0.032] \\
Gemini 2.5 Pro & 200 & +0.014 [0.001, 0.029] & -0.074 [-0.100, -0.048] \\
Claude Opus 4.7 & 200 & -0.012 [-0.019, -0.006] & -0.005 [-0.016, 0.006] \\
GPT-5.4 & 200 & -0.001 [-0.016, 0.012] & -0.004 [-0.012, 0.004] \\
DeepSeek V3.2 & 145 & -0.028 [-0.045, -0.012] & +0.043 [0.019, 0.068] \\
GPT-4o & 200 & -0.054 [-0.074, -0.034] & -0.007 [-0.020, 0.006] \\
Claude 4.5 Haiku & 200 & -0.006 [-0.020, 0.007] & +0.041 [0.020, 0.063] \\
Llama 4 Maverick & 190 & -0.010 [-0.027, 0.007] & -0.003 [-0.028, 0.022] \\
Qwen3-32B & 187 & -0.011 [-0.032, 0.010] & -0.010 [-0.030, 0.011] \\
\bottomrule
\end{tabular}
\end{table}

\subsection{Withholding mechanism definitions}

Our benchmark provides candidate mechanism definitions alongside NLL feedback, enabling agents to formulate mechanistic hypotheses and use them to guide experiment selection. To examine how access to these definitions affects
identification accuracy and exploration behavior, we conducted an exploratory ablation in which the detailed mechanism definitions were withheld.
Gemini 3.1 Pro and Gemini 2.5 Pro received candidate IDs M1-M20 and per-candidate NLL feedback, while retaining the general catalytic substrate-to-product task description.

\paragraph{Prompt without the detailed mechanism definitions}
\begin{PromptBox}
You are an expert in chemical kinetics and experimental design.
Your mission is to hypothesize plausible mechanisms, and design a single, maximally informative experiment to help distinguish between the plausible candidate mechanisms.
You may propose {num_proposals} new experiments.

Execute the following same steps iteratively.
At the beginning of each turn, before executing any step, check how many iterations are done based on the conversation history.

1. **Hypothesize about plausible mechanisms**
    *   You are provided with the fitting results unless this is the first turn. Based on the fitting results, identify all mechanisms that still have non-negligible support.
    *   Even if some mechanisms appear less likely, do NOT prematurely commit to a single leading hypothesis.
    *   Treat hypothesis diversity as important information to preserve.
2. **Hypothesize optimal experiment for discrimination**
    *   Your goal is NOT to maximize the distinction between the single most likely pair of mechanisms.
    *   Instead, your goal is to reduce uncertainty over the entire set of remaining plausible mechanisms.
    *   Prefer experiments that are informative for many mechanisms simultaneously.
    *   Avoid experiments that would only distinguish the top 1–2 candidates while leaving the others indistinguishable.
    *   Consider how the proposed condition helps differentiate multiple mechanisms at once and why it avoids overly myopic or extreme designs.
    *   Reflect on the history of conditions: avoid repeating similar regimes, but also avoid extreme conditions that collapse mechanistic differences (e.g., saturation, depletion).
3.  **Propose Experiment**:
    *   The experimental conditions that you can propose are initial concentration of substrate, catalyst and product.
    *   Conclude with <suggestion> ... </suggestion> and <<DONE>>.

After you propose a new experiment, you would obtain the fitting result of all runs including new experimental data based on your suggestion.

For each turn, you must follow this format:
<iteration>
Specify the iteration number. Start from 1 and increment by 1 iteration.
</iteration>
<reflection>
- Your observations and reasoning about the previous turn's results.
- A clear statement on whether your aim for the turn was achieved.
</reflection>
<hypothesis>
Your detailed reasoning for the proposed experiment. This should include your scientific hypothesis, how the experiment will distinguish between mechanisms, and why the design is effective.
</hypothesis>
<candidate_mechanisms>
List the plausible candidate mechanisms, sorted in descending order of plausibility (most likely first). Provide them as a JSON list of strings (e.g. ["M1", "M2"]).
</candidate_mechanisms>
<suggestion>
{{
    "initial_concentration_of_substrate": <float 0.1 - 5.0>,
    "initial_concentration_of_catalyst": <float 5e-4 - 0.20>,
    "initial_concentration_of_product": <float 0.0 - 0.5>,
}}
</suggestion>
<<DONE>>

# 20 Candidate Mechanisms
There are 20 candidate mechanisms, denoted M1 to M20.
All 20 candidates correspond to distinct possible catalytic mechanisms for the conversion of substrate S to product P using catalyst cat.
However, the detailed mechanistic definitions of M1–M20 are hidden from you.

# Data Structure
The following describes the structure of the data. This is not provided to you directly, but you can access the fitting results of all runs through the conversation history.
At the first run, there is only "run_1". As you propose new experiments and obtain new data, additional runs (e.g., "run_2", "run_3") will be added.

```python
{{
  "run_1": {{
    "initial_concentration_of_catalyst": float, # Example: 0.01
    "time_data": list[float], # List of time points
    "product_data": list[float], # List of product concentrations at each time point
    "substrate_data": list[float] # List of substrate concentrations at each time point
  }},
  "run_2": {{
    # ... same structure as run_1
  }},
  # ... additional runs
}}
```

# Fitting Results
Everytime you propose a new experiment, you will receive the fitting results for all runs including the new experimental data based on your suggestion.
The fitting results are provided as negative log-likelihood values for each candidate mechanism.
Here are example fitting results from the previous runs:
```json
{{
    "M1": -66.4739509769402,
    "M2": -66.18465035678079,
    "M3": -118.16423286937294,
    "M4": -131.86593439082938,
    "M5": 35.36495497243471,
    "M6": -78.6224862051261,
    "M7": -77.53127516407199,
    "M8": 69.72349746655837,
    "M9": -93.85551875223946,
    "M10": -106.0115552520685,
    "M11": -111.90660321199617,
    "M12": -117.2823521990672,
    "M13": -127.68391782844185,
    "M14": -66.05561653283398,
    "M15": -135.0936662137469,
    "M16": -71.82102553018892,
    "M17": -66.07488048869106,
    "M18": -63.6419421066368,
    "M19": -88.57669991706247,
    "M20": -48.13563459951533
}}
```
\end{PromptBox}

Table \ref{tab:mechanism_definition_ablation} summarizes identification accuracy and exploration behavior. Gemini 3.1 Pro showed little change in point-estimate accuracy when mechanism definitions were withheld (0.610 to 0.605), remaining above the LHS baseline (0.465) and BO baseline (0.425) point estimates. In contrast, Gemini 2.5 Pro's accuracy decreased from 0.575 to 0.440. Both models exhibited higher occupancy entropy without mechanism definitions, indicating broader coverage of the discretized experimental space. Their corner-probing behavior differed: S-cat corner mass increased for Gemini 3.1 Pro (0.256 to 0.284), whereas it decreased for Gemini 2.5 Pro (0.175 to 0.105).

\begin{table}[h]
\centering
\caption{
Effect of withholding candidate mechanism definitions.
Changes are computed as definitions withheld minus definitions provided.
}
\label{tab:mechanism_definition_ablation}
\small
\setlength{\tabcolsep}{4pt}
\begin{tabular}{llccc}
\hline
Model & Metric &
\shortstack{Definitions\\provided} &
\shortstack{Definitions\\withheld} &
Change \\
\hline
Gemini 3.1 Pro & Accuracy
& 0.610 & 0.605 & -0.005 \\
Gemini 2.5 Pro & Accuracy
& 0.575 & 0.440 & -0.135 \\
\hline
Gemini 3.1 Pro & Occupancy entropy
& 0.492 & 0.521 & +0.028 \\
Gemini 2.5 Pro & Occupancy entropy
& 0.471 & 0.522 & +0.052 \\
\hline
Gemini 3.1 Pro & S--cat corner mass
& 0.256 & 0.284 & +0.028 \\
Gemini 2.5 Pro & S--cat corner mass
& 0.175 & 0.105 & -0.070 \\
\hline
\end{tabular}
\end{table}

Withholding mechanism definitions reduced accuracy for Gemini 2.5 Pro, but not uniformly across the two models. These results do not isolate the contribution of mechanism-specific hypothesis formation to task performance. Withholding definitions removes the explicit mechanistic bias for connecting candidate differences to target interventions, while leaving general kinetic knowledge and feedback-driven exploration available. 
Both models changed their exploration behavior when definitions were withheld, suggesting a shift in experiment-selection strategy. Such a shift may have helped Gemini 3.1 Pro retain a similar accuracy point estimate, whereas Gemini 2.5 Pro's point estimate declined. Similar point estimates do not establish equivalence, and both models are from the same provider.
These results suggest that agents may respond to missing mechanism definitions by adopting alternative exploration strategies, with differing success in maintaining accuracy. Consequently, the accuracy changes in this ablation alone cannot distinguish the contribution of reasoning based on the supplied mechanism definitions from that of alternative exploration strategies.

\section{Compute resources}
\label{app:compute}

\subsection{Compute}
For each LLM agent and baseline, the full benchmark requires fitting 20 mechanisms $\times$ 5 iterations $\times$ 200 systems = 20{,}000 parameter-estimation runs (CMA-ES, 60 seconds per fit timeout; see Appendix \ref{app:fitting}). Parallelized over 70 CPU cores, this took approximately 5 hours of wall-clock time per model ($\sim$350 CPU-hours). Across the 10 evaluated LLM models and 3 baselines, total fitting compute was approximately $\sim$4,550 CPU-hours.

\subsection{LLM API usage}
LLM agent inference was performed via the OpenRouter API. Total API cost across all 10 evaluated agents $\times$ 200 systems $\times$ up to 5 iterations was on the order of \$50 (frontier models such as Gemini 3 Pro and GPT-5.4 account for the majority of this cost; open-weight models served via OpenRouter are substantially cheaper). The original iteration-3 LLM-as-a-judge evaluation (400 traces $\times$ 4 rubrics scored by GPT-5.4 and Claude Opus 4.7) cost approximately \$40; this estimate excludes the additional iteration and prompt-ablation evaluations.

\section{Representative reasoning traces}
\label{app:reasoning_trace_examples}

\begin{table}[h]
\centering
\caption{Example of a high-scoring reasoning trace. The scores for all the steps, which are belief update, hypothesis discrimination, the scores for discriminative experiment planning and predictive reasoning are 2.}
\small
\setlength{\tabcolsep}{5pt}
\renewcommand{\arraystretch}{1.15}
\begin{tabularx}{\linewidth}{p{0.24\linewidth}X}
\toprule
\multicolumn{2}{l}{\textbf{Strong reasoning trace: Gemini 3.1 Pro}} \\
\midrule
\\

\textbf{Context}
&
The NLLs identify M7, M8, and M6 as the leading mechanisms, with M3 as a secondary candidate. The main uncertainty is whether the reaction is driven by a substrate-assisted complex (M7), catalyst activation (M6/M8), or catalyst dimerization (M3).
\\

\textbf{Belief update}
&
The trace correctly updates the belief state: M7 is treated as the current leader, while M6 and M8 remain close alternatives. M3 is kept as a lower-probability but mechanistically relevant candidate.
\\

\textbf{Hypothesis discrimination}
&
The trace identifies the key contrast: M7 needs formation of a substrate-rich complex, catS2, whereas M6/M8 rely on activated catalyst, cat*. M7 should therefore be especially sensitive to whether enough substrate is available for a second substrate-binding step, while M6/M8 should depend more on catalyst activation.
\\

\textbf{Experiment planning}
&
The proposed experiment uses a substrate-starvation regime, $[\mathrm{S}]_0=0.15$ and $[\mathrm{cat}]_0=0.15$. This directly targets the M7 versus M6/M8 distinction: if M7 is correct, the first substrate-binding step consumes substrate and leaves little free substrate for catS2 formation; if M6/M8 are correct, catalyst activation can still proceed.
\\

\textbf{Predictive reasoning}
&
The trace gives mechanism-specific predictions: under substrate starvation, M7 should stall or slow sharply because catS2 formation requires an additional substrate molecule, whereas M6/M8 should retain activity through cat* formation. This maps the proposed condition to concrete, distinguishable kinetic outcomes.
\\
\bottomrule
\end{tabularx}
\label{tab:reasoning_trace_strong}
\end{table}

\begin{table}[h]
\centering
\caption{Example of a lower-scoring reasoning trace. The scores for belief update and hypothesis discrimination is 2, and the scores for discriminative experiment planning and predictive reasoning is 1.}
\small
\setlength{\tabcolsep}{5pt}
\renewcommand{\arraystretch}{1.15}
\begin{tabularx}{\linewidth}{p{0.24\linewidth}X}
\toprule
\multicolumn{2}{l}{\textbf{Weak planning/prediction trace: Mistral Large 3 on M6}} \\
\midrule

\textbf{Context}
&
The NLLs support a very similar hypothesis space: M6 and M8 are the top candidates, with M7 and M3 still plausible. Thus both traces reason over essentially the same mechanism families.
\\

\textbf{Belief update}
&
The trace tracks the NLL feedback: it identifies M6 and M8 as dominant, with M7 and M3 still plausible but less likely.
\\

\textbf{Hypothesis discrimination}
&
The trace correctly states that M6 and M8 differ in activation reversibility: M6 has irreversible activation, whereas M8 has reversible activation with ligand release/rebinding. It also recognizes M7 and M3 as alternative mechanism classes.
\\

\textbf{Experiment planning}
&
The proposed experiment uses $[\mathrm{S}]_0=1.0$, $[\mathrm{cat}]_0=0.02$, and $[\mathrm{P}]_0=0.2$. The trace states that product is added to test whether the system can ``reset'' under reversible activation in M8, while the catalyst increase is intended to amplify activation effects. However, product addition is not directly tied to the main M6/M8 activation-reversibility contrast, and the modest catalyst increase is not a sharp catalyst-scaling test.
\\

\textbf{Predictive reasoning}
&
The trace predicts that product may affect M8 but not M6, and that the higher catalyst concentration may amplify activation effects. These claims are directionally related to the stated hypotheses, but the trace does not specify a clear observable signature that would separate M6 from M8, nor does it give concrete predictions for M7 or M3 under the proposed condition.
\\
\bottomrule
\end{tabularx}
\label{tab:reasoning_trace_weak}
\end{table}

\clearpage
\section*{NeurIPS Paper Checklist}

\begin{enumerate}

\item {\bf Claims}
    \item[] Question: Do the main claims made in the abstract and introduction accurately reflect the paper's contributions and scope?
    \item[] Answer: \answerYes{} %
    \item[] Justification: The abstract and introduction clearly state the paper's main contributions. These claims are directly supported by the experimental results and analysis presented in Section \ref{sec:experiments}. The scope and limitations of the claims, including the closed-world setting and task-specific nature of the reasoning decomposition, are explicitly discussed in Section \ref{sec:limitation}.

    \item[] Guidelines:
    \begin{itemize}
        \item The answer \answerNA{} means that the abstract and introduction do not include the claims made in the paper.
        \item The abstract and/or introduction should clearly state the claims made, including the contributions made in the paper and important assumptions and limitations. A \answerNo{} or \answerNA{} answer to this question will not be perceived well by the reviewers. 
        \item The claims made should match theoretical and experimental results, and reflect how much the results can be expected to generalize to other settings. 
        \item It is fine to include aspirational goals as motivation as long as it is clear that these goals are not attained by the paper. 
    \end{itemize}

\item {\bf Limitations}
    \item[] Question: Does the paper discuss the limitations of the work performed by the authors?
    \item[] Answer: \answerYes{}{} %
    \item[] Justification: The limitations of the proposed benchmark are discussed in Section \ref{sec:limitation}, including the closed-world system identification setting and its implications for generalization and hypothesis generation.
    \item[] Guidelines:
    \begin{itemize}
        \item The answer \answerNA{} means that the paper has no limitation while the answer \answerNo{} means that the paper has limitations, but those are not discussed in the paper. 
        \item The authors are encouraged to create a separate ``Limitations'' section in their paper.
        \item The paper should point out any strong assumptions and how robust the results are to violations of these assumptions (e.g., independence assumptions, noiseless settings, model well-specification, asymptotic approximations only holding locally). The authors should reflect on how these assumptions might be violated in practice and what the implications would be.
        \item The authors should reflect on the scope of the claims made, e.g., if the approach was only tested on a few datasets or with a few runs. In general, empirical results often depend on implicit assumptions, which should be articulated.
        \item The authors should reflect on the factors that influence the performance of the approach. For example, a facial recognition algorithm may perform poorly when image resolution is low or images are taken in low lighting. Or a speech-to-text system might not be used reliably to provide closed captions for online lectures because it fails to handle technical jargon.
        \item The authors should discuss the computational efficiency of the proposed algorithms and how they scale with dataset size.
        \item If applicable, the authors should discuss possible limitations of their approach to address problems of privacy and fairness.
        \item While the authors might fear that complete honesty about limitations might be used by reviewers as grounds for rejection, a worse outcome might be that reviewers discover limitations that aren't acknowledged in the paper. The authors should use their best judgment and recognize that individual actions in favor of transparency play an important role in developing norms that preserve the integrity of the community. Reviewers will be specifically instructed to not penalize honesty concerning limitations.
    \end{itemize}

\item {\bf Theory assumptions and proofs}
    \item[] Question: For each theoretical result, does the paper provide the full set of assumptions and a complete (and correct) proof?
    \item[] Answer: \answerNA{} %
    \item[] Justification: The paper does not present theoretical results such as theorems or formal guarantees requiring assumptions and proofs. The contributions are empirical and methodological, focusing on benchmark design and experimental evaluation.
    \item[] Guidelines:
    \begin{itemize}
        \item The answer \answerNA{} means that the paper does not include theoretical results. 
        \item All the theorems, formulas, and proofs in the paper should be numbered and cross-referenced.
        \item All assumptions should be clearly stated or referenced in the statement of any theorems.
        \item The proofs can either appear in the main paper or the supplemental material, but if they appear in the supplemental material, the authors are encouraged to provide a short proof sketch to provide intuition. 
        \item Inversely, any informal proof provided in the core of the paper should be complemented by formal proofs provided in appendix or supplemental material.
        \item Theorems and Lemmas that the proof relies upon should be properly referenced. 
    \end{itemize}

    \item {\bf Experimental result reproducibility}
    \item[] Question: Does the paper fully disclose all the information needed to reproduce the main experimental results of the paper to the extent that it affects the main claims and/or conclusions of the paper (regardless of whether the code and data are provided or not)?
    \item[] Answer: \answerYes{} %
    \item[] Justification: The benchmark setup (200 reaction systems, 20 candidate mechanisms) is described in Section \ref{sec:benchmark_overview} with mechanism definitions in Appendix \ref{app:candidate_mechanisms}. The simulator details (COPASI configuration, observation model, noise level) are in Appendix \ref{app:simulator_data_generation}. Baselines (Maximin LHS, Bayesian optimization, Tool-based EIG reference) are described in Section \ref{sec:baselines}. The 10 evaluated LLM agents and inference configurations (temperature, max tokens, timeout) are listed in Appendix \ref{app:agent_prompt}. The code URL is specified in the abstract.
    \item[] Guidelines:
    \begin{itemize}
        \item The answer \answerNA{} means that the paper does not include experiments.
        \item If the paper includes experiments, a \answerNo{} answer to this question will not be perceived well by the reviewers: Making the paper reproducible is important, regardless of whether the code and data are provided or not.
        \item If the contribution is a dataset and\slash or model, the authors should describe the steps taken to make their results reproducible or verifiable. 
        \item Depending on the contribution, reproducibility can be accomplished in various ways. For example, if the contribution is a novel architecture, describing the architecture fully might suffice, or if the contribution is a specific model and empirical evaluation, it may be necessary to either make it possible for others to replicate the model with the same dataset, or provide access to the model. In general. releasing code and data is often one good way to accomplish this, but reproducibility can also be provided via detailed instructions for how to replicate the results, access to a hosted model (e.g., in the case of a large language model), releasing of a model checkpoint, or other means that are appropriate to the research performed.
        \item While NeurIPS does not require releasing code, the conference does require all submissions to provide some reasonable avenue for reproducibility, which may depend on the nature of the contribution. For example
        \begin{enumerate}
            \item If the contribution is primarily a new algorithm, the paper should make it clear how to reproduce that algorithm.
            \item If the contribution is primarily a new model architecture, the paper should describe the architecture clearly and fully.
            \item If the contribution is a new model (e.g., a large language model), then there should either be a way to access this model for reproducing the results or a way to reproduce the model (e.g., with an open-source dataset or instructions for how to construct the dataset).
            \item We recognize that reproducibility may be tricky in some cases, in which case authors are welcome to describe the particular way they provide for reproducibility. In the case of closed-source models, it may be that access to the model is limited in some way (e.g., to registered users), but it should be possible for other researchers to have some path to reproducing or verifying the results.
        \end{enumerate}
    \end{itemize}

\item {\bf Open access to data and code}
    \item[] Question: Does the paper provide open access to the data and code, with sufficient instructions to faithfully reproduce the main experimental results, as described in supplemental material?
    \item[] Answer: \answerYes{} %
    \item[] Justification: This paper provides open access to the data and code, and include instructions for running the code. We include the links to our code at the end of the abstract.
    \item[] Guidelines:
    \begin{itemize}
        \item The answer \answerNA{} means that paper does not include experiments requiring code.
        \item Please see the NeurIPS code and data submission guidelines (\url{https://neurips.cc/public/guides/CodeSubmissionPolicy}) for more details.
        \item While we encourage the release of code and data, we understand that this might not be possible, so \answerNo{} is an acceptable answer. Papers cannot be rejected simply for not including code, unless this is central to the contribution (e.g., for a new open-source benchmark).
        \item The instructions should contain the exact command and environment needed to run to reproduce the results. See the NeurIPS code and data submission guidelines (\url{https://neurips.cc/public/guides/CodeSubmissionPolicy}) for more details.
        \item The authors should provide instructions on data access and preparation, including how to access the raw data, preprocessed data, intermediate data, and generated data, etc.
        \item The authors should provide scripts to reproduce all experimental results for the new proposed method and baselines. If only a subset of experiments are reproducible, they should state which ones are omitted from the script and why.
        \item At submission time, to preserve anonymity, the authors should release anonymized versions (if applicable).
        \item Providing as much information as possible in supplemental material (appended to the paper) is recommended, but including URLs to data and code is permitted.
    \end{itemize}

\item {\bf Experimental setting/details}
    \item[] Question: Does the paper specify all the training and test details (e.g., data splits, hyperparameters, how they were chosen, type of optimizer) necessary to understand the results?
    \item[] Answer: \answerYes{} %
    \item[] Justification: Benchmark overview, benchmark size, the 20 mechanisms definitions, and observation noise are given in Section \ref{sec:benchmark_overview} and Appendix \ref{app:simulator_data_generation}. Details and hyperparameters of Maximin LHS baseline, Bayes baseline and EIG reference are specified in Section \ref{app:baselines}. LLM configurations are specified in Appendix \ref{app:agent_prompt}.
    \item[] Guidelines:
    \begin{itemize}
        \item The answer \answerNA{} means that the paper does not include experiments.
        \item The experimental setting should be presented in the core of the paper to a level of detail that is necessary to appreciate the results and make sense of them.
        \item The full details can be provided either with the code, in appendix, or as supplemental material.
    \end{itemize}

\item {\bf Experiment statistical significance}
    \item[] Question: Does the paper report error bars suitably and correctly defined or other appropriate information about the statistical significance of the experiments?
    \item[] Answer: \answerYes{} %
    \item[] Justification: All error bars in the experimental section are 95 \% bootstrap confidence intervals. The bootstrap procedure is specified in Section \ref{sec:experiments}
    \item[] Guidelines:
    \begin{itemize}
        \item The answer \answerNA{} means that the paper does not include experiments.
        \item The authors should answer \answerYes{} if the results are accompanied by error bars, confidence intervals, or statistical significance tests, at least for the experiments that support the main claims of the paper.
        \item The factors of variability that the error bars are capturing should be clearly stated (for example, train/test split, initialization, random drawing of some parameter, or overall run with given experimental conditions).
        \item The method for calculating the error bars should be explained (closed form formula, call to a library function, bootstrap, etc.)
        \item The assumptions made should be given (e.g., Normally distributed errors).
        \item It should be clear whether the error bar is the standard deviation or the standard error of the mean.
        \item It is OK to report 1-sigma error bars, but one should state it. The authors should preferably report a 2-sigma error bar than state that they have a 96\% CI, if the hypothesis of Normality of errors is not verified.
        \item For asymmetric distributions, the authors should be careful not to show in tables or figures symmetric error bars that would yield results that are out of range (e.g., negative error rates).
        \item If error bars are reported in tables or plots, the authors should explain in the text how they were calculated and reference the corresponding figures or tables in the text.
    \end{itemize}

\item {\bf Experiments compute resources}
    \item[] Question: For each experiment, does the paper provide sufficient information on the computer resources (type of compute workers, memory, time of execution) needed to reproduce the experiments?
    \item[] Answer: \answerYes{} %
    \item[] Justification: CPU usage and LLM API costs, are reported in Appendix \ref{app:compute}.
    \item[] Guidelines:
    \begin{itemize}
        \item The answer \answerNA{} means that the paper does not include experiments.
        \item The paper should indicate the type of compute workers CPU or GPU, internal cluster, or cloud provider, including relevant memory and storage.
        \item The paper should provide the amount of compute required for each of the individual experimental runs as well as estimate the total compute. 
        \item The paper should disclose whether the full research project required more compute than the experiments reported in the paper (e.g., preliminary or failed experiments that didn't make it into the paper). 
    \end{itemize}
    
\item {\bf Code of ethics}
    \item[] Question: Does the research conducted in the paper conform, in every respect, with the NeurIPS Code of Ethics \url{https://neurips.cc/public/EthicsGuidelines}?
    \item[] Answer: \answerYes{} %
    \item[] Justification: The research conforms with the NeurIPS Code of Ethics. The benchmark uses entirely synthetic kinetic systems; no human subjects, personal data, or sensitive content are involved. LLM agents are evaluated on a scientific reasoning task.
    \item[] Guidelines:
    \begin{itemize}
        \item The answer \answerNA{} means that the authors have not reviewed the NeurIPS Code of Ethics.
        \item If the authors answer \answerNo, they should explain the special circumstances that require a deviation from the Code of Ethics.
        \item The authors should make sure to preserve anonymity (e.g., if there is a special consideration due to laws or regulations in their jurisdiction).
    \end{itemize}

\item {\bf Broader impacts}
    \item[] Question: Does the paper discuss both potential positive societal impacts and negative societal impacts of the work performed?
    \item[] Answer: \answerYes{} %
    \item[] Justification: Potential benefits are discussed in Section \ref{sec:conclusion}. Section \ref{sec:limitation} and the hosted Croissant metadata discuss synthetic-data limitations and the risk of overgeneralizing benchmark performance to wet-lab deployment.
    
    \item[] Guidelines:
    \begin{itemize}
        \item The answer \answerNA{} means that there is no societal impact of the work performed.
        \item If the authors answer \answerNA{} or \answerNo, they should explain why their work has no societal impact or why the paper does not address societal impact.
        \item Examples of negative societal impacts include potential malicious or unintended uses (e.g., disinformation, generating fake profiles, surveillance), fairness considerations (e.g., deployment of technologies that could make decisions that unfairly impact specific groups), privacy considerations, and security considerations.
        \item The conference expects that many papers will be foundational research and not tied to particular applications, let alone deployments. However, if there is a direct path to any negative applications, the authors should point it out. For example, it is legitimate to point out that an improvement in the quality of generative models could be used to generate Deepfakes for disinformation. On the other hand, it is not needed to point out that a generic algorithm for optimizing neural networks could enable people to train models that generate Deepfakes faster.
        \item The authors should consider possible harms that could arise when the technology is being used as intended and functioning correctly, harms that could arise when the technology is being used as intended but gives incorrect results, and harms following from (intentional or unintentional) misuse of the technology.
        \item If there are negative societal impacts, the authors could also discuss possible mitigation strategies (e.g., gated release of models, providing defenses in addition to attacks, mechanisms for monitoring misuse, mechanisms to monitor how a system learns from feedback over time, improving the efficiency and accessibility of ML).
    \end{itemize}
    
\item {\bf Safeguards}
    \item[] Question: Does the paper describe safeguards that have been put in place for responsible release of data or models that have a high risk for misuse (e.g., pre-trained language models, image generators, or scraped datasets)?
    \item[] Answer: \answerNA{} %
    \item[] Justification: The paper does not release pre-trained models, generative image/text models, or scraped datasets. The released artifacts consist of synthetic kinetic simulation data, evaluation code, and rubric prompts, none of which present a meaningful risk of misuse.
    \item[] Guidelines:
    \begin{itemize}
        \item The answer \answerNA{} means that the paper poses no such risks.
        \item Released models that have a high risk for misuse or dual-use should be released with necessary safeguards to allow for controlled use of the model, for example by requiring that users adhere to usage guidelines or restrictions to access the model or implementing safety filters. 
        \item Datasets that have been scraped from the Internet could pose safety risks. The authors should describe how they avoided releasing unsafe images.
        \item We recognize that providing effective safeguards is challenging, and many papers do not require this, but we encourage authors to take this into account and make a best faith effort.
    \end{itemize}

\item {\bf Licenses for existing assets}
    \item[] Question: Are the creators or original owners of assets (e.g., code, data, models), used in the paper, properly credited and are the license and terms of use explicitly mentioned and properly respected?
    \item[] Answer: \answerYes{} %
    \item[] Justification: Third-party assets used in the paper are cited in the references.
    \item[] Guidelines:
    \begin{itemize}
        \item The answer \answerNA{} means that the paper does not use existing assets.
        \item The authors should cite the original paper that produced the code package or dataset.
        \item The authors should state which version of the asset is used and, if possible, include a URL.
        \item The name of the license (e.g., CC-BY 4.0) should be included for each asset.
        \item For scraped data from a particular source (e.g., website), the copyright and terms of service of that source should be provided.
        \item If assets are released, the license, copyright information, and terms of use in the package should be provided. For popular datasets, \url{paperswithcode.com/datasets} has curated licenses for some datasets. Their licensing guide can help determine the license of a dataset.
        \item For existing datasets that are re-packaged, both the original license and the license of the derived asset (if it has changed) should be provided.
        \item If this information is not available online, the authors are encouraged to reach out to the asset's creators.
    \end{itemize}

\item {\bf New assets}
    \item[] Question: Are new assets introduced in the paper well documented and is the documentation provided alongside the assets?
    \item[] Answer: \answerYes{} %
    \item[] Justification: We introduce a new benchmark, an evaluation pipeline, and a reasoning-decomposition rubric with LLM-as-a-judge prompts. All new assets are documented in Sections \ref{sec:benchmark_overview} - \ref{sec:reasoning_analysis}. The release includes a README documenting code structure, licensing, and reproduction instructions. No human-subject data is involved.
    \item[] Guidelines:
    \begin{itemize}
        \item The answer \answerNA{} means that the paper does not release new assets.
        \item Researchers should communicate the details of the dataset\slash code\slash model as part of their submissions via structured templates. This includes details about training, license, limitations, etc. 
        \item The paper should discuss whether and how consent was obtained from people whose asset is used.
        \item At submission time, remember to anonymize your assets (if applicable). You can either create an anonymized URL or include an anonymized zip file.
    \end{itemize}

\item {\bf Crowdsourcing and research with human subjects}
    \item[] Question: For crowdsourcing experiments and research with human subjects, does the paper include the full text of instructions given to participants and screenshots, if applicable, as well as details about compensation (if any)? 
    \item[] Answer: \answerNA{} %
    \item[] Justification: Human input comprises author calibration on 30 reasoning traces and independent annotations by two researchers with chemistry expertise on 60 traces (Appendix \ref{app:judge_agreement}). The two expert annotators were labmates who volunteered without compensation. The rubric is provided in the appendix; the released code includes the annotation instructions and interface. This was a voluntary assessment of model outputs, rather than a participant study.
    \item[] Guidelines:
    \begin{itemize}
        \item The answer \answerNA{} means that the paper does not involve crowdsourcing nor research with human subjects.
        \item Including this information in the supplemental material is fine, but if the main contribution of the paper involves human subjects, then as much detail as possible should be included in the main paper. 
        \item According to the NeurIPS Code of Ethics, workers involved in data collection, curation, or other labor should be paid at least the minimum wage in the country of the data collector. 
    \end{itemize}

\item {\bf Institutional review board (IRB) approvals or equivalent for research with human subjects}
    \item[] Question: Does the paper describe potential risks incurred by study participants, whether such risks were disclosed to the subjects, and whether Institutional Review Board (IRB) approvals (or an equivalent approval/review based on the requirements of your country or institution) were obtained?
    \item[] Answer: \answerNA{} %
    \item[] Justification: The authors considered formal ethics review unnecessary for the voluntary assessment of model outputs by two unpaid labmates; no formal review was obtained. The annotation activity and its scope are described in Appendix \ref{app:judge_agreement}.
    \item[] Guidelines:
    \begin{itemize}
        \item The answer \answerNA{} means that the paper does not involve crowdsourcing nor research with human subjects.
        \item Depending on the country in which research is conducted, IRB approval (or equivalent) may be required for any human subjects research. If you obtained IRB approval, you should clearly state this in the paper. 
        \item We recognize that the procedures for this may vary significantly between institutions and locations, and we expect authors to adhere to the NeurIPS Code of Ethics and the guidelines for their institution. 
        \item For initial submissions, do not include any information that would break anonymity (if applicable), such as the institution conducting the review.
    \end{itemize}

\item {\bf Declaration of LLM usage}
    \item[] Question: Does the paper describe the usage of LLMs if it is an important, original, or non-standard component of the core methods in this research? Note that if the LLM is used only for writing, editing, or formatting purposes and does \emph{not} impact the core methodology, scientific rigor, or originality of the research, declaration is not required.
    \item[] Answer: \answerYes{} %
    \item[] Justification: LLMs are central methodological components of this work in two  roles. (i) As evaluation subjects: 10 LLM agents are evaluated as scientific reasoning agents on our benchmark. This is written in Section \ref{sec:experiments} (ii) As automated rubric judges. This is written in Section \ref{sec:reasoning_analysis}.
    \item[] Guidelines:
    \begin{itemize}
        \item The answer \answerNA{} means that the core method development in this research does not involve LLMs as any important, original, or non-standard components.
        \item Please refer to our LLM policy in the NeurIPS handbook for what should or should not be described.
    \end{itemize}

\end{enumerate}

\end{document}